\documentclass[11pt,a4paper]{article}
\usepackage[hyperref]{emnlp-ijcnlp-2019}
\usepackage{times}
\usepackage{latexsym}
\usepackage{microtype}

\usepackage{tikz}
\usepackage{tikz-dependency}
\usepackage{booktabs}
\usepackage{subcaption}
\usepackage{placeins} 
\usepackage{colortbl}

\usepackage{amsmath}
\usepackage{amssymb}

\usepackage[disable]{todonotes}
\usepackage{comment}

\usepackage{url}
\usepackage{cleveref}

\aclfinalcopy 

\newcommand\blfootnote[1]{%
  \begingroup
  \renewcommand\thefootnote{}\footnote{#1}%
  \addtocounter{footnote}{-1}%
  \endgroup
}

\title{Linking artificial and human neural representations of language}

\author{Jon Gauthier \and Roger P. Levy \\
  Massachusetts Institute of Technology \\
  Department of Brain and Cognitive Sciences \\
  {\tt jon@gauthiers.net, rplevy@mit.edu}}

\date{}

\begin{document}
\maketitle
\begin{abstract}
What information from an act of sentence understanding is robustly represented in the human brain? We investigate this question by comparing sentence encoding models on a brain decoding task, where the sentence that an experimental participant has seen must be predicted from the fMRI signal evoked by the sentence. We take a pre-trained BERT architecture as a baseline sentence encoding model and fine-tune it on a variety of natural language understanding (NLU) tasks, asking which lead to improvements in brain-decoding performance.

We find that none of the sentence encoding tasks tested yield significant increases in brain decoding performance. Through further task ablations and representational analyses, we find that tasks which produce syntax-light representations yield significant improvements in brain decoding performance. Our results constrain the space of NLU models that could best account for human neural representations of language, but also suggest limits on the possibility of decoding fine-grained syntactic information from fMRI human neuroimaging.
\end{abstract}
\blfootnote{\noindent{}Source code for all analyses reported in this paper is available at \url{http://bit.ly/nn-decoding}.}

What are the neural representations which support human language understanding? Language neuroscience has converged on a set of reliable physiological markers related to language processing \citep{kutas2011thirty}, and a picture of where in the brain different aspects of language processing take place \citep{fedorenko2010new}. But we still largely lack a coherent picture of the \emph{structure} and \emph{format} of the neural representations driving language understanding.

Part of this struggle is due to a scarcity of candidate representational structures in the first place. While there are certainly enough representational theories of language understanding, many are specified at too high a level of analysis to plausibly map onto neural structures without serious further revision \citep{poeppel2012maps}.

Studies which draw on these high-level representations must therefore also assume some link between such representations and measures of neural activity --- for example, that the magnitude of neural activations should match up with the probability values derived from a computational model \citep{brennan2016abstract}, or measures derived from syntactic representations of the input \citep{pallier2011cortical}. While the success of these mapping studies demonstrates that specific summary statistics of linguistic representations have correlates in the mind, these summary statistics are not themselves candidate representations for the fundamental operations underlying language understanding.

In the meantime, research in natural language processing has produced neural network models that capture many different sorts of intelligent language understanding behavior \citep{collobert2011natural,goldberg2016primer}. These models accomplish this behavior in an implementation better matched with that of the brain, with information about their inputs distributed across a high-dimensional continuous space. Could these models be taken seriously as candidate hypotheses of how language processing could be implemented in neural hardware? Under the assumption that both human brain and neural network representations are optimally suited to solve some \emph{task} \citep{anderson1990adaptive}, linking these two computational systems should reveal parallel task-optimal structure shared within their representations.

This paradigm linking brain and machine has already seen substantial success in vision science. \citet{yamins2013hierarchical} first demonstrated that the activations of a convolutional neural network trained on ImageNet in response to natural images could predict activations in a macaque monkey's visual cortex in response to the same images. This result and others have
led to an increasingly detailed understanding of the contents of brain representations \citep{schrimpf2018brain} and novel artificial neural network architectures \citep{kubilius2018cornet} in the domain of vision.

In language understanding, several authors have exploited neural network representations as proxies for sentence meaning, and demonstrated that human brain activations in response to sentences can match with these meaning representations at well above chance performance \citep[see e.g.][]{mitchell2008predicting,wehbe2014aligning,huth2016natural,pereira2018toward}. Our aim in this paper is to further understand why these mappings are successful, uncovering the parallel \emph{representational contents} shared between human brains and neural networks.

We evaluate the link between human brain activity and neural network models as the models are optimized for different tasks. We find that neural network models quickly diverge in their capacity to match human brain activations as they are optimized for different NLU objectives. We further locate correlates of these changes in representational content, finding that the granularity of a model's \emph{syntactic} representations is at least partially responsible for their differences in brain decoding performance. Overall, this approach allows us to generate and validate hypotheses about the representational contents of both human brain and neural network activity.

\section{Related work}

Several papers have begun to explore the brain--machine link in language understanding, asking whether human brain activations can be matched with the activations of computational language models.
\citet{mitchell2008predicting} first demonstrated that distributional word representations could be used to predict human brain activations, when subjects were presented with individual words in isolation. \citet{huth2016natural} replicated and extended these results using distributed word representations, and \citet{pereira2018toward} extended these results to sentence stimuli. \citet{wehbe2014aligning}, \citet{qian2016bridging}, \citet{jain2018incorporating}, and \citet{abnar2019blackbox} next introduced more complex word and sentence meaning representations, demonstrating that neural network language models could better account for brain activation by incorporating representations of longer-term linguistic context. \citet{gauthier2018word} and \citet{sun2019towards} further demonstrated that optimizing model representations for different objectives yielded substantial differences in brain decoding performance.
This paper extends the neural network brain decoding paradigm both in \textbf{breadth}, studying a wide class of different task-optimal models, and in \textbf{depth}, exploring the particular representational contents of each model responsible for its brain decoding performance.

\section{Methods}
\label{sec:methods}

\begin{figure}[t]
    \centering
    \includegraphics[width=0.85\linewidth]{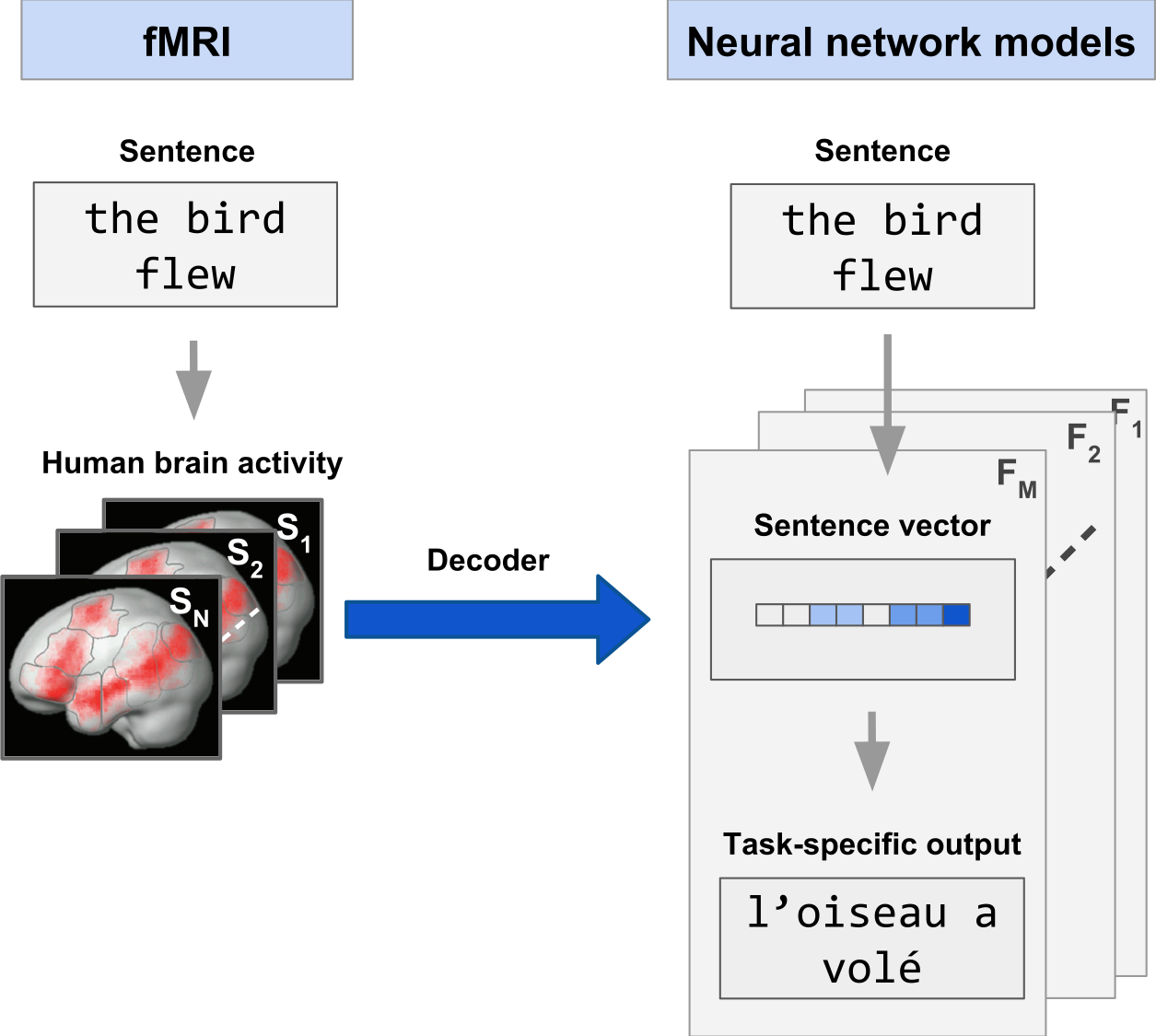}
    \caption{Brain decoding methodology. We use human brain activations in response to sentences to predict how neural networks represent those same sentences.}
    \label{fig:method}
\end{figure}

\Cref{fig:method} describes the high-level design of our experiments, which attempt to match human neuroimaging data with different candidate model representations of sentence inputs. Using a dataset of human brain activations recorded in response to complete sentences, we learn linear regression models which map from human brain activity to representations of the same sentences produced by different natural language understanding models.

To the extent that this linear decoding is successful, it can reveal parallel structure between brain and model representations. Consider a softmax neural network classifier model optimized for some task $T$, mapping input sentences $\mathbf x$ to class outputs $y$. We can factor this neural network classifier into the composition of two operations, a representational function $r(\mathbf x)$, and an affine operator $A$:
\begin{equation}
    p(y \mid \mathbf x) \propto \exp({A(r(\mathbf x))})
\end{equation}
Under this factorization, $r(\mathbf x)$ must compute representations which are linearly separable with respect to the classes of the task $T$.

Research in cognitive neuroscience has shown that surprisingly many features of perceptual and cognitive states are likewise linearly separable from images of human brain activity, even at the coarse spatial and temporal resolution afforded by functional magnetic resonance imaging \citep[fMRI; see e.g.][]{haxby2001distributed,kriegeskorte2006information}. However, the full power of linear decoding with fMRI remains unknown within language neuroscience and elsewhere. One possibility (1) is that the representational distinctions intrinsically required to describe language understanding behavior are linearly decodable from fMRI data. If this were the case, we could use performance in brain decoding to gauge the similarity between the mental representations underlying human language understanding and those deployed within artificial neural network models. Conversely (2), if the representations supporting language understanding in the brain are not linearly decodable from fMRI, we should be able to demonstrate this fact by showing specific ablations of sentence representation models do  not degrade in brain decoding performance. Thus, the brain decoding framework offers possibilities both for (1) discriminating among NLU tasks as faithful characterizations of human language understanding, and for (2) understanding potential limitations of fMRI imaging and linear decoding methods. We explore both of these possibilities in this paper.


\Cref{sec:methods-neuroimaging} describes the human neuroimaging data used as the source of this learned mapping. \Cref{sec:methods-models} next describes how we derive the target representations of sentence inputs from different natural language understanding models. Finally, \Cref{sec:methods-decoding} describes our method for deriving and evaluating mappings between the two representational spaces.

\subsection{Human neuroimaging data}
\label{sec:methods-neuroimaging}
\newcommand\npereira{384}

\begin{table}[t]
    \centering
    \resizebox{\linewidth}{!}{
    \begin{tabular}{p{12cm}}
        \toprule
        A man is a male human adult. \\
        Fingers are used for grasping, writing, grooming and other activities. \\
        Piranhas have very sharp teeth and are opportunistic carnivores. \\
        Pressing a piano key causes a felt-tipped hammer to hit a vibrating steel string. \\
        A range of mountains forms due to tectonic plate collision. \\
        \bottomrule
    \end{tabular}
    }
    \caption{Randomly sampled sentence stimuli from the human neuroimaging dataset of \citet{pereira2018toward}.}
    \label{tbl:pereira-sentences}
\end{table}

We use the human brain images collected by \citet[experiment 2]{pereira2018toward}.\footnote{This dataset is publicly available at \url{https://osf.io/crwz7/}.} \citeauthor{pereira2018toward} visually presented \npereira{} natural-language sentences to 8 adult subjects. The sentences (examples in \Cref{tbl:pereira-sentences}) consisted of simple encyclopedic statements about natural kinds, written by the authors. The subjects were instructed to carefully read each sentence, presented one at a time, and think about its meaning. As they read the sentences, the subjects' brain activity was recorded with functional magnetic resonance imaging (fMRI). For each subject and each sentence, the fMRI images consist of a $\sim 200,000$-dimensional vector describing the approximate neural activity within small 3D patches of the brain, known as voxels.\footnote{For more information on data collection and fMRI analysis, please see \citet{pereira2018toward}.} We collect these vectors in a single matrix and compress them to $d_B = 256$ dimensions using PCA, yielding a matrix $B_i \in \mathbb R^{\npereira{}\times d_B}$ for each subject $i$.\footnote{The PCA projections retain more than 95\% of the variance among sentence responses within each subject.}

\subsection{Sentence representation models}
\label{sec:methods-models}
\newcommand\thetalm{\ensuremath{{\Theta_\text{LM}}}}
\newcommand\thetaf{\ensuremath{{\Theta_{j\ell}}}}
\newcommand\csubj{\ensuremath{B_i}}
\newcommand\clm{\ensuremath{C_\text{LM}}}
\newcommand\cf{\ensuremath{C_{j\ell}}}
\newcommand\decoder{\ensuremath{G_{i\to{}j\ell}}}
\newcommand\decoderloss{\ensuremath{J_{ij\ell}}}

\begin{figure*}[t]
\begin{subfigure}{0.245\linewidth}
\includegraphics[width=\linewidth]{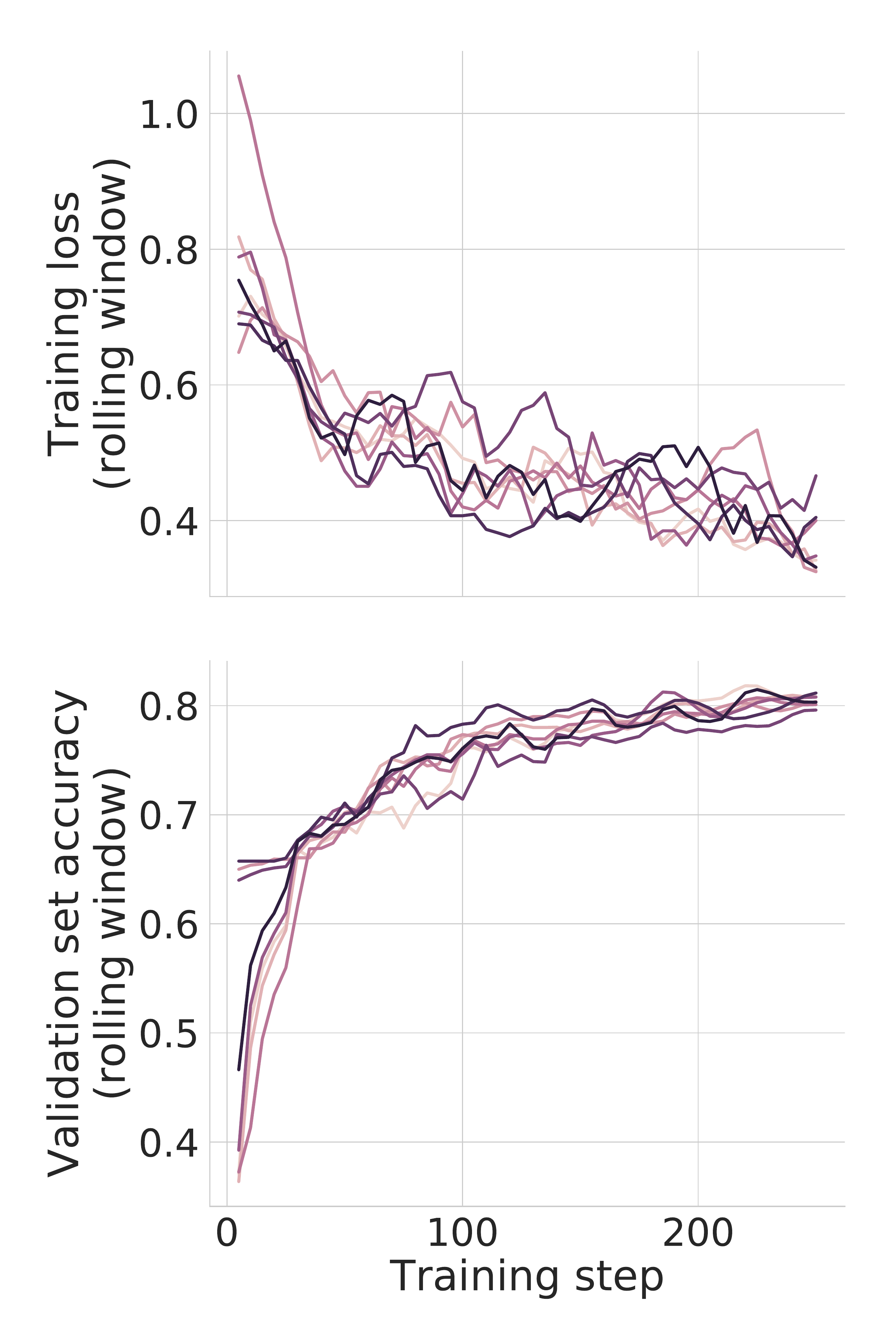}
\caption{QQP}
\end{subfigure}\hfill%
\begin{subfigure}{0.245\linewidth}
\includegraphics[width=\linewidth]{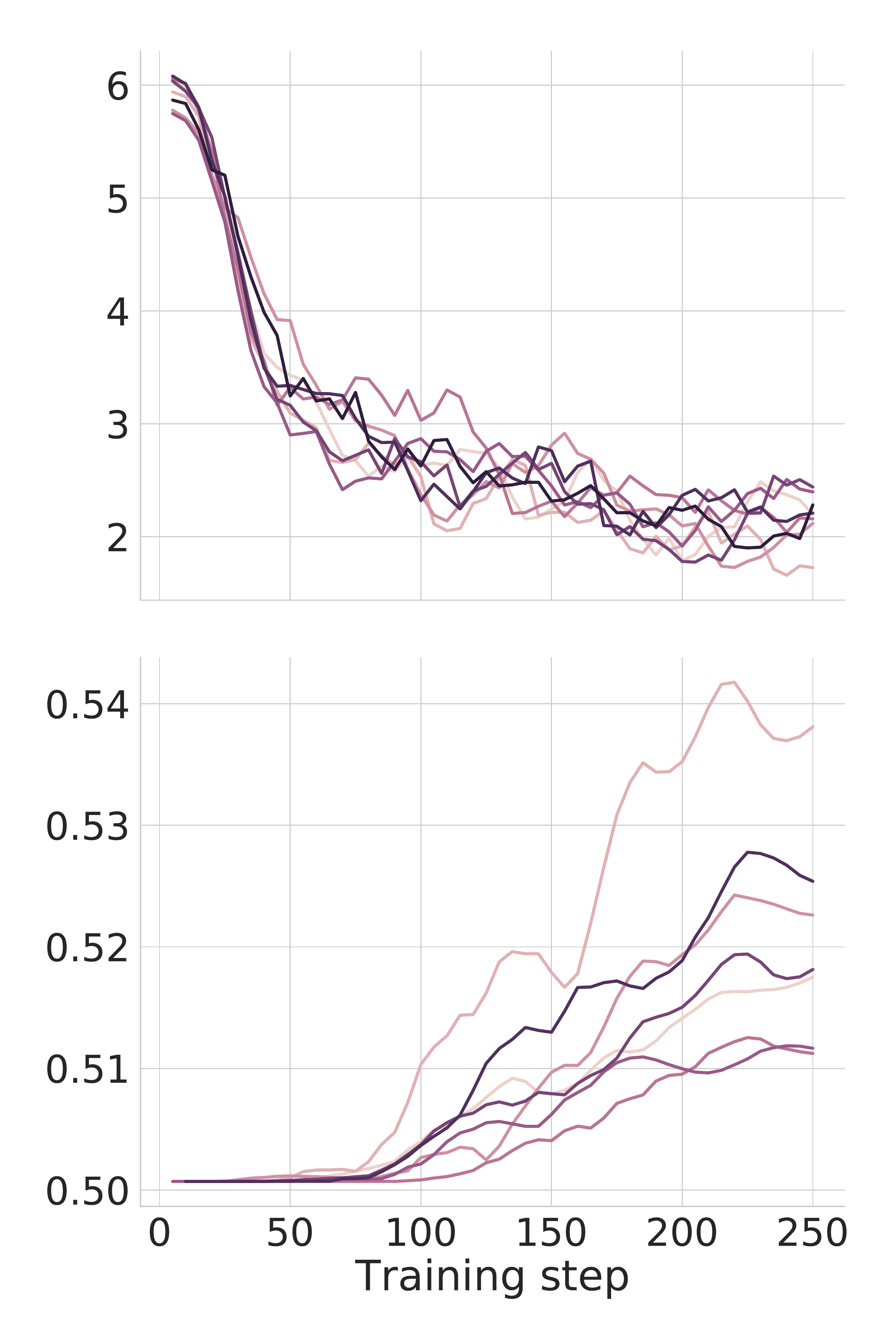}
\caption{SQuAD}
\end{subfigure}\hfill%
\begin{subfigure}{0.245\linewidth}
\includegraphics[width=\linewidth]{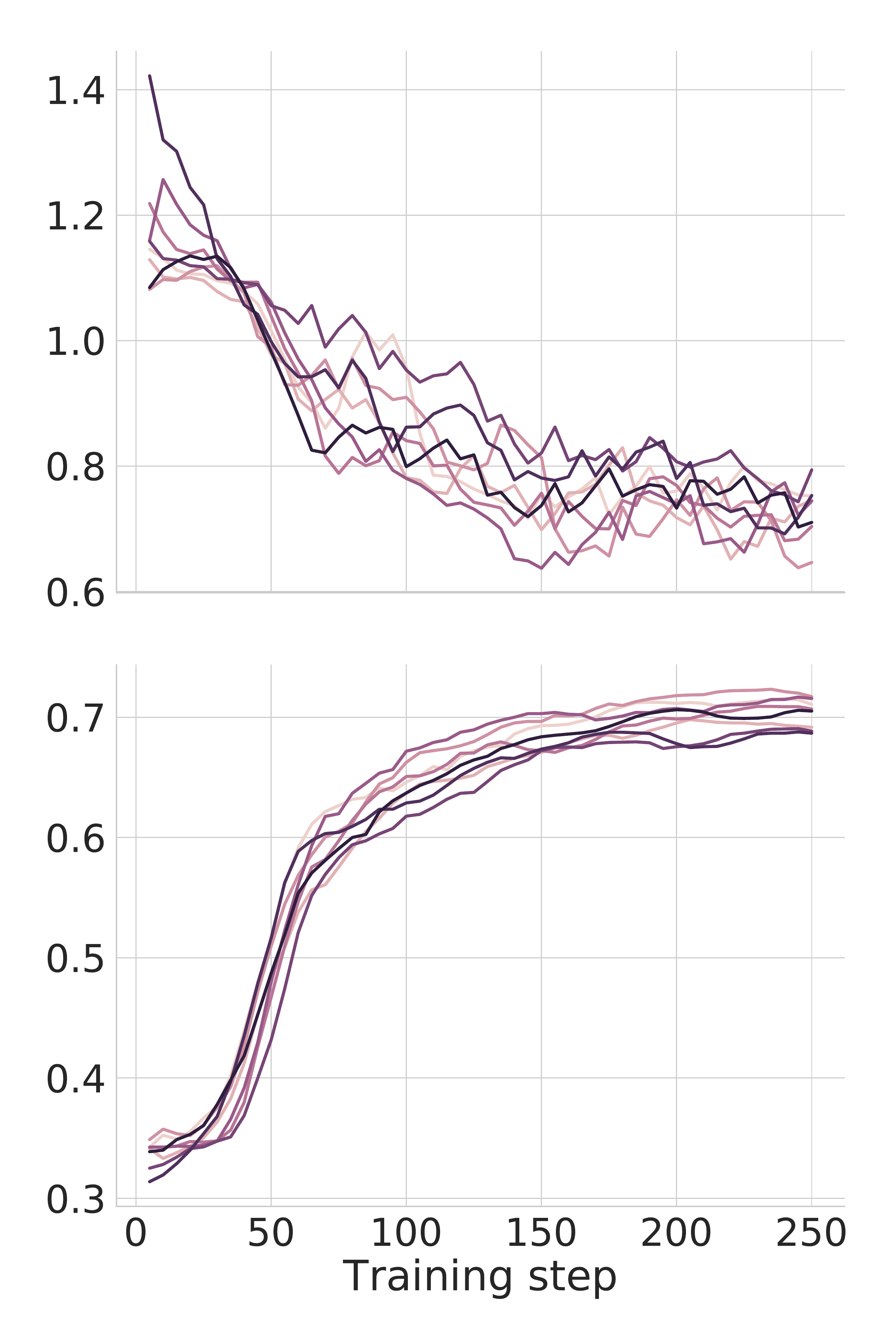}
\caption{MNLI}
\end{subfigure}\hfill%
\begin{subfigure}{0.245\linewidth}
\includegraphics[width=\linewidth]{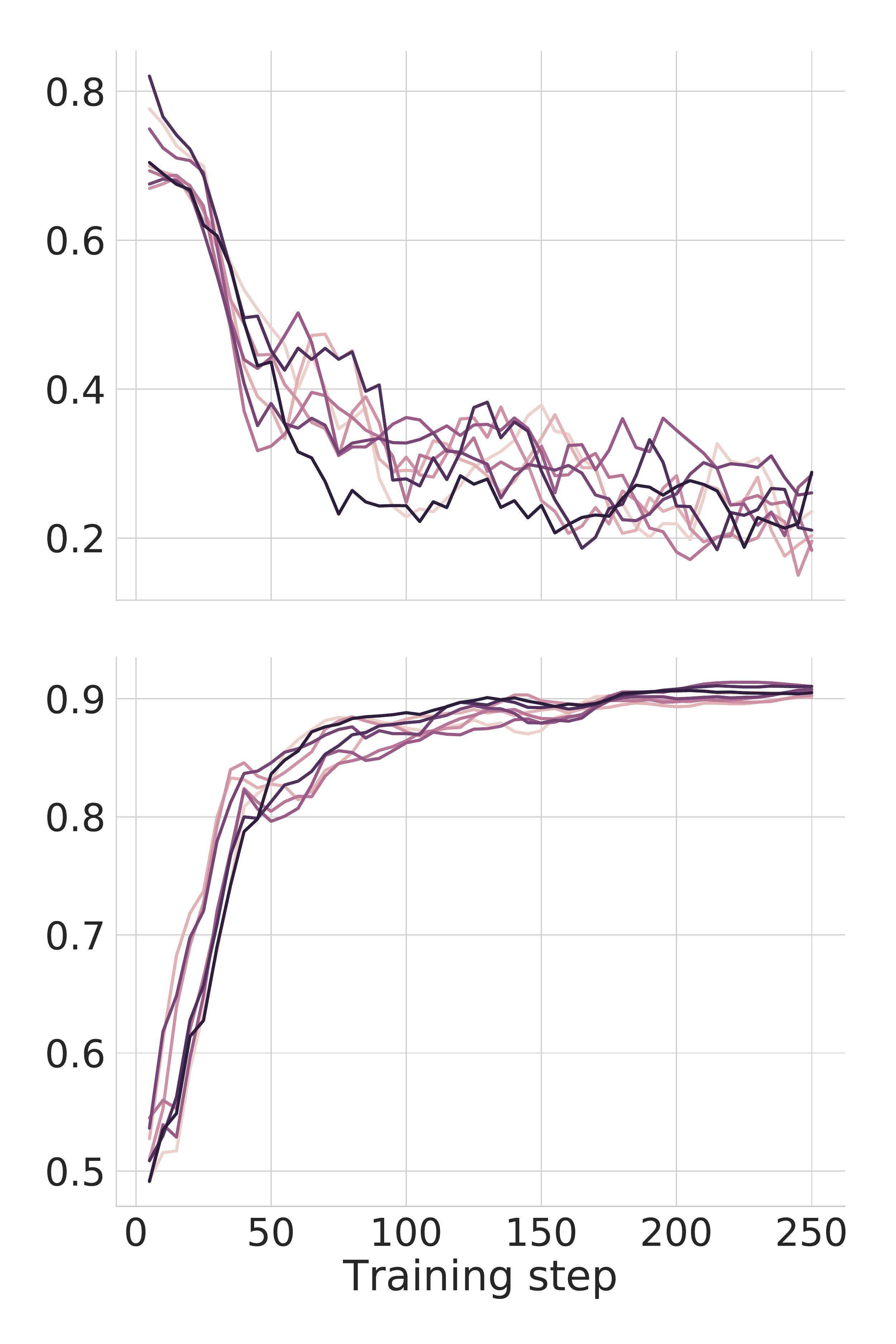}
\caption{SST}
\end{subfigure}
\caption{Learning curves (training set loss and evaluation set accuracy) for NLU model fine-tuning.}
\label{fig:learning-curves}
\end{figure*}

\begin{table}[t]
    \centering
    \resizebox{\linewidth}{!}{
    \begin{tabular}{p{1.5cm}p{2cm}p{1.3cm}|ccc}
        \toprule
        Task & Dataset & Domain & \# train & Avg sent len. & \# types \\
        \midrule
        Paraphrase classification & Quora Question Pairs & social QA & 364k & 22.1 & 103k \\ 
        \arrayrulecolor{black!30}\midrule
        Question answering & SQuAD 2.0 \citep{rajpurkar2018know} & wiki & 130k & 11.2 & 43.9k \\
        \midrule
        Natural language inference & MNLI \citep{williams2017broad} & mixed & 393k & 16.8 & 83.3k \\
        \midrule
        Sentiment analysis & SST-2 \citep{socher2013recursive} & movie reviews & 67.3k & 9.41 & 14.8k \\
        \arrayrulecolor{black}\bottomrule
    \end{tabular}
    }
    \caption{Details of the tasks used for fine-tuning.}
    \label{tbl:fine-tuning-tasks}
\end{table}

We will match the human brain activation data described above with a suite of different sentence representations. Our primary concern in this evaluation is to compare alternative \emph{tasks} and the representational contents they demand, rather than comparing neural network architectures. For this reason, we draw sentence representations from a unified neural network architecture --- the bidirectional Transformer model BERT \citep{devlin2018bert} --- as we optimize it to perform different tasks.

The BERT model uses a series of multi-head attention operations to compute context-sensitive representations for each token in an input sentence. The model is pre-trained on two tasks: (1) a cloze language modeling task, where the model is given a complete sentence containing several masked words and asked to predict the identity of a particular masked word; and (2) a next-sentence prediction task, where the model is given two sentences and asked to predict whether the sentences are immediately adjacent in the original language modeling data.\footnote{See \citet{devlin2018bert,vaswani2017attention} for further details on model architecture and training.} For our purposes, this pre-training process produces a set of BERT parameters $\thetalm$ jointly optimized for these two objectives, consisting of word embeddings and attention mechanism parameters.

For an input token sequence $w_1,\ldots,w_T$, the output of the BERT model is a corresponding sequence of contextualized representations of each token. We derive a single sentence representation vector by prepending a constant token $w_0 = \texttt{[CLS]}$, and extracting its corresponding output vector at the final layer, following \citet{devlin2018bert}.\footnote{We also repeated the experiments of this paper using sentence representations computed by uniformly averaging BERT's contextualized token representations of each input sentence, and found similar qualitative results as will be presented below.}

We first extract sentence representations from the pre-trained model parameterized by \thetalm. We let $\clm \in \mathbb R^{384 \times d_H}$ refer to the matrix of sentence representations drawn from this pre-trained BERT model, where $d_H$ is the dimensionality of the BERT hidden layer representation.

\subsubsection{Fine-tuned models}
We use the code and pre-trained weights released by \citet{devlin2018bert} to fine-tune the BERT model on a suite of natural language processing classification tasks, shown in \Cref{tbl:fine-tuning-tasks}.\footnote{Details on the fine-tuning procedure are available in \citet{devlin2018bert}.}
This fine-tuning process jointly optimizes the pre-trained word embeddings and attention weights drawn from $\thetalm$, along with a task-specific classification model which accepts the sentence representations produced by the BERT model (the vector corresponding to the prepended \texttt{[CLS]} token) as input.

We fine-tune the pretrained BERT model on a set of popular shared NLU tasks, shown in \Cref{tbl:fine-tuning-tasks}, with fixed hyperparameters across tasks (available in \Cref{sec:appendix-hyperparameters}). Each fine-tuning operation is run for 250 iterations, before which all models show substantial improvements on the fine-tuning task. \Cref{fig:learning-curves} shows the learning curves for each of the models fine-tuned by this procedure.

We execute 8 different runs for each fine-tuning task. Each run $\ell$ of each fine-tuning task $j$ produces a set of final parameters $\thetaf$. We can derive a set of sentence representations $\cf$ from each fine-tuned model by the same logic as above, feeding in each sentence with a prepended \texttt{[CLS]} token and extracting the contextualized \texttt{[CLS]} representation.

For our purposes, the product of each fine-tuning run is a set of sentence representations $\cf \in \mathbb R^{\npereira \times d_H}$.

\paragraph{Custom tasks}

In order to better understand \emph{why} models might fail or succeed at brain decoding, we also produced several custom fine-tuning tasks. Each task was a modified form of the standard BERT cloze language modeling task, manipulated to strongly select for or against some particular aspect of linguistic representation.\footnote{The training and test data for these tasks were generated from the Toronto Books Corpus \citep{zhu2015aligning}, a subset of the data used to train the original BERT model. Further details and examples for these tasks are given in \Cref{sec:appendix-custom-tasks}.}

\subparagraph{Scrambled language modeling} We first design two language modeling tasks to select against fine-grained syntactic representation of inputs. We randomly shuffle words from the corpus samples used for language modeling, to remove all first-order cues to syntactic structure. Our first custom task, \texttt{LM-scrambled}, deals with sentence inputs where words are shuffled within sentences; our second task, \texttt{LM-scrambled-para}, uses inputs where words are shuffled within their containing paragraphs in the corpus. \footnote{This shuffling method removes first-order cues to constituency: for example, the fact that \emph{the table} appears directly to the right of \emph{on} suggests that they are members of a single constituent. It is still possible that the model can exploit second-order cues to structure. For example, if two nouns \emph{ranch} and \emph{rancher} appear in the same sentence as \emph{dressing}, we can still guess that \emph{ranch} is more likely to modify \emph{dressing}.}

By shuffling inputs in this way, we effectively turn the cloze task into a bag-of-words language modeling task: given a set of words from a sentence or a random draw of words from a paragraph, the model must predict a missing word.
After optimizing models on these scrambled tasks, we design a probe to validate the effects of the task on the model's syntactic representations. This probe is detailed in \Cref{sec:syntactic-probe}.

\subparagraph{Part-of-speech language modeling} We next design a task, \texttt{LM-pos}, to select against fine-grained semantic representation of inputs. We do this by requiring a model to predict only the \emph{part of speech} of a masked word, rather than the word itself. This manipulation removes pressure for the model to distinguish predictions between target words in the same syntactic class.

\subparagraph{} We repeat fine-tuning runs on each of these custom tasks 4 times per task, and see substantial improvements in held-out task performance for each of these custom tasks after just 250 steps of training. 
After fine-tuning, we extract sentence representation matrices $\cf$ from each run $\ell$ of each these models.

\paragraph{Language modeling control} As a control, we also continue training on the original BERT language modeling objectives using text drawn from the Books Corpus \citep{zhu2015aligning}. We run 4 fine-tuning runs of this task, yielding representations $C_{\text{LM},\ell}$.

\subsubsection{Word vector baseline}

As a baseline comparison, we also include sentence representations computed from GloVe word vectors \citep{pennington2014glove}. Unlike BERT's word representations, these word vectors are insensitive to their surrounding sentential context. These word vectors have nevertheless successfully served as sentence meaning representations in prior studies \citep{pereira2018toward,gauthier2018word}. We let $C_\text{GloVe}(w_1,\ldots,w_T) = \frac 1 T e(w_t)$, where $e(w_i)$ retrieves the GloVe embedding for word $w_i$.\footnote{We use publicly available GloVe vectors computed on Common Crawl, available in the spaCy toolkit as \texttt{en\textunderscore{}vectors\textunderscore{}web\textunderscore{}lg}.}

\subsection{Brain decoding}
\label{sec:methods-decoding}

We next learn a suite of \emph{decoders}: regression models mapping from descriptions of human brain activation to model activations in response to sentences.
Let $\csubj \in \mathbb R^{384 \times d_B}$ represent the brain activations of subject $i$ in response to the 384 sentences in our evaluation set. For each subject and each sentence representation available (including both language model representations \clm{} and fine-tuned model representations \cf), we learn a linear map $\decoder : \mathbb R^{d_H \times d_B}$ between the two spaces which minimizes the regularized regression loss
\begin{equation}
    \decoderloss = ||\decoder \csubj - \cf||_2^2 + \beta\, ||\decoder||_2^2
\end{equation}
\noindent{}where $\beta$ is a regularization hyperparameter. For each subject's collection of brain images and each target model representation, we train and evaluate the above regression model with nested 8-fold cross-validation \citep{cawley2010over}. The regression models are evaluated under two metrics: mean squared error (MSE) in prediction of model activations, and average rank (AR):
    \begin{equation}
      \text{AR}(i, j, \ell) = \frac{1}{384} \sum_k \text{rank}((\decoder \csubj)[k], \cf[k])
    \end{equation}
    where $\text{rank}((\decoder \csubj)[k], \cf[k])$ gives the rank of a ground-truth sentence representation $\cf[k]$ in the list of nearest neighbors of a predicted sentence representation $(\decoder \csubj)[k]$, ordered by increasing cosine distance.
    
These two metrics serve complementary roles: the MSE metric strictly evaluates the ability of human brain activations to exactly match the representational geometry of model activations, while the AR metric simply requires that the brain activations be able to support the relevant meaning contrasts between the \npereira{} sentences tested.

\section{Results}

\begin{figure*}[t]
    \centering
    \hfill%
    \begin{subfigure}{0.49\linewidth}
    \includegraphics[width=\linewidth]{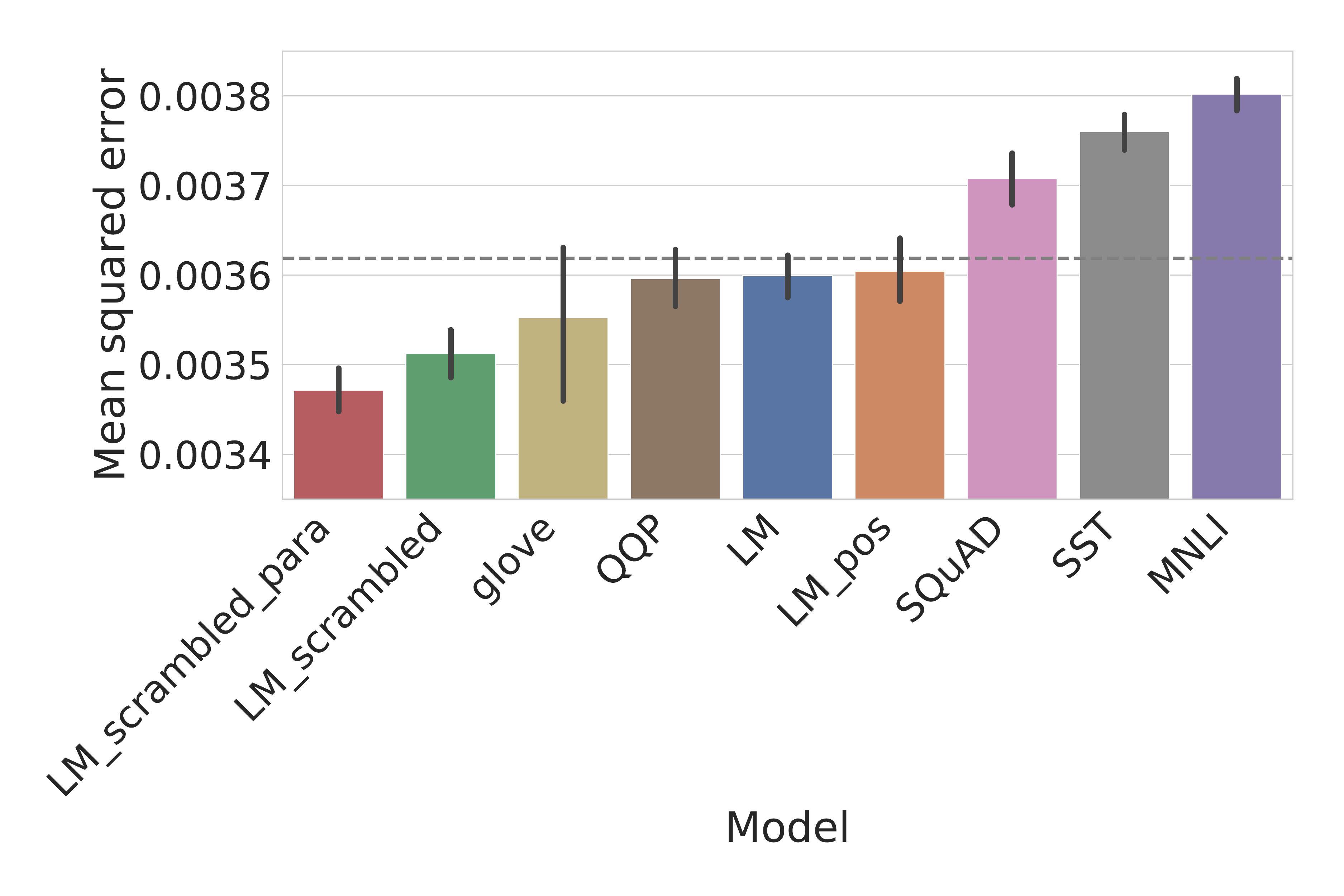}
    \caption{Mean squared error metric}
    \end{subfigure}\hfill%
    \begin{subfigure}{0.49\linewidth}
    \includegraphics[width=\linewidth]{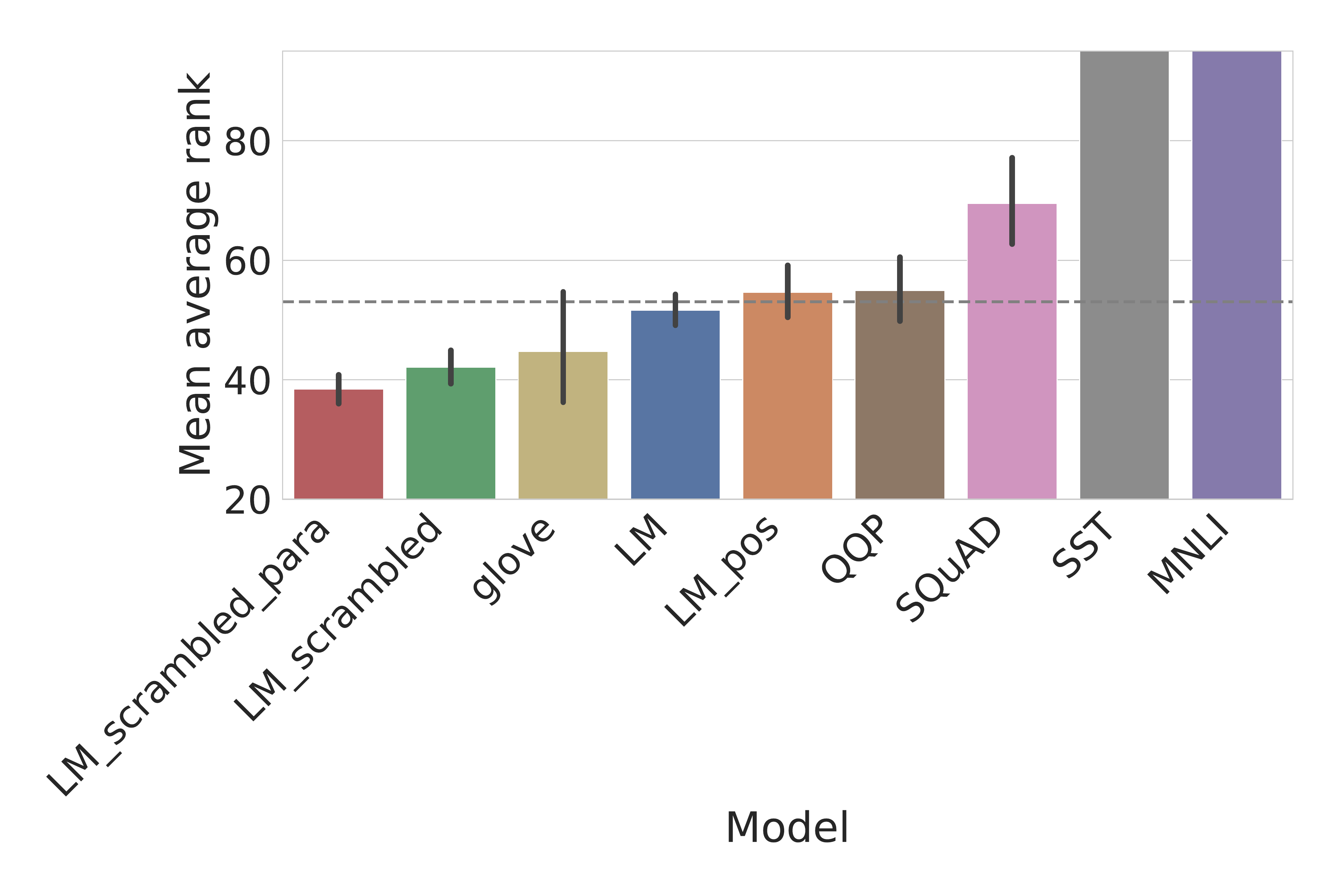}
    \caption{Average rank metric}
    \end{subfigure}%
    \hfill
    \caption{Brain decoding performance for fine-tuned BERT models and the GloVe baseline. Error bars are bootstrapped 95\% confidence intervals, pooling across decoders matching different subjects (8 total) and different BERT fine-tuning runs (up to 8 runs per model). Dashed gray line shows the average brain decoding performance of the pretrained BERT language model.}
    \label{fig:decoding-final}
\end{figure*}

We first present the performance of all of the BERT models tested and the GloVe baseline in \Cref{fig:decoding-final}. This figure makes apparent a number of surprising findings, which we validate by paired $t$-tests.\footnote{Each sample in our statistical tests compares the brain decoding performance matching a subject's brain image with model representations before and after fine-tuning on a particular task. See \Cref{fig:decoding-final-within-subject,fig:decoding-final-within-model} of \Cref{sec:appendix-supp-fig} for further visualizations. Results are reported throughout with a significance level $\alpha = 0.01$.}

\begin{enumerate}

    \item On average, fine-tuning on the standard NLU tasks yields \emph{increased error} in brain decoding under both metrics relative to the BERT baseline (MSE, $t \approx 14.8, p < 3 \times 10^{-36}$; AR, $t \approx 17, p < 3 \times 10^{-42}$). This trend is significant for each model individually, except QQP (MSE, $t \approx -2.2, p > 0.03$; AR, $t \approx 0.79, p > 0.4$).
    \item Fine-tuning on the LM-scrambled-para custom task yields \emph{decreased error} in brain decoding under both metrics relative to the BERT baseline (MSE, $t \approx -33.6, p < 3 \times 10^{-45}$; AR, $t \approx -26.9, p < 5 \times 10^{-39}$) and GloVe vectors (MSE, $t \approx -23.4, p < 4 \times 10^{-35}$; AR, $t \approx -12.1, p < 6 \times 10^{-19})$.
    \item Fine-tuning on both the control language-modeling task and the LM-pos custom task yields ambiguous results: \emph{decreased MSE} and \emph{no significant change in AR} relative to the BERT baseline (LM-pos: MSE, $t \approx -2.90, p < 0.007$; AR, $t \approx 2.10, p > 0.04$; LM: MSE, $t \approx -5.68, p < 4 \times 10^{-7}$; AR, $t \approx -2.36, p > 0.02$).
\end{enumerate}


\subsection{Learning dynamics}

\begin{figure*}[t]
    \centering
    \begin{subfigure}{0.499\linewidth}
    \includegraphics[width=\linewidth]{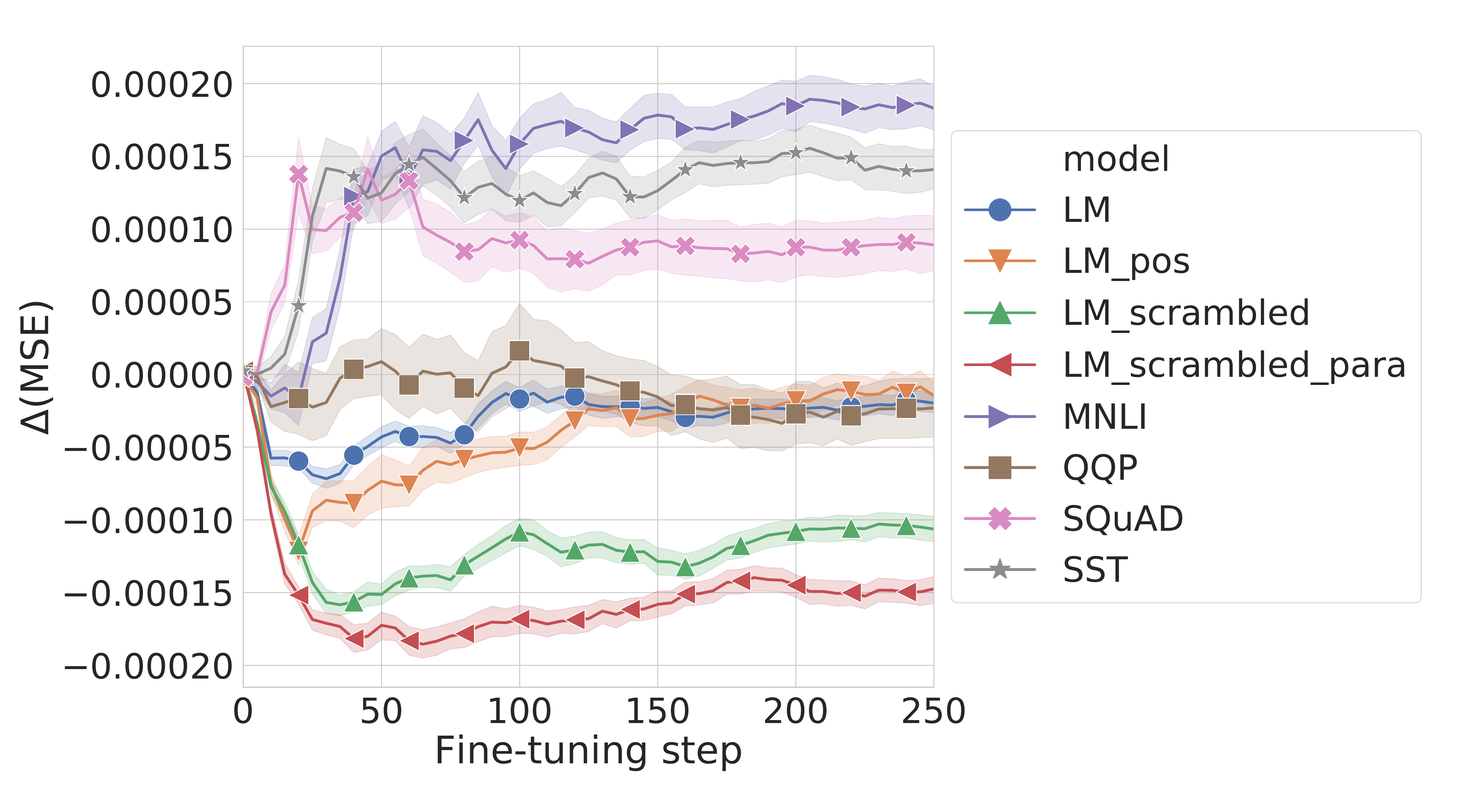}
    \caption{Mean squared-error metric}
    \end{subfigure}\hfill%
    \begin{subfigure}{0.499\linewidth}
    \includegraphics[width=\linewidth]{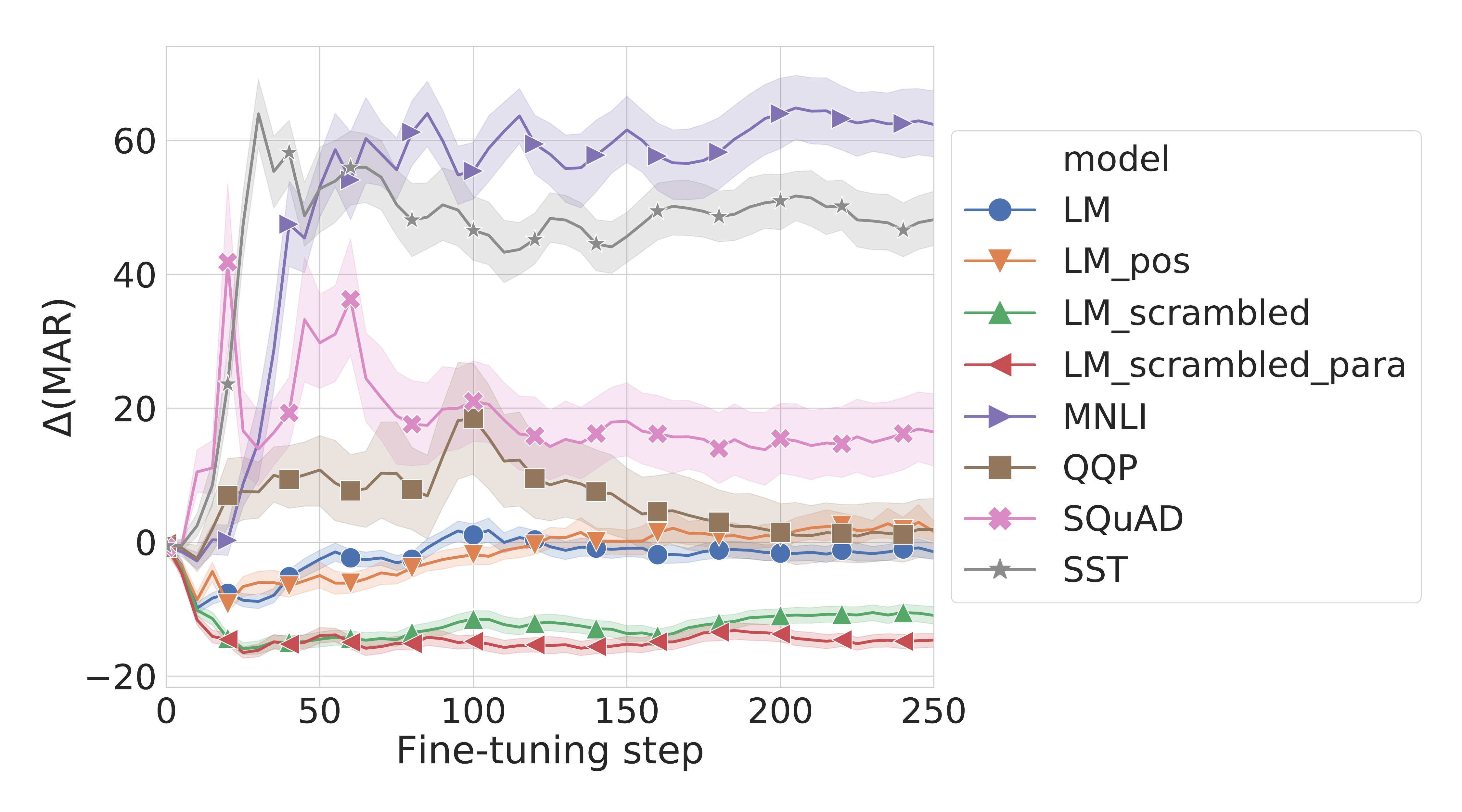}
    \caption{Average rank metric}
    \end{subfigure}
    \caption{Brain decoding performance trajectories over fine-tuning time, graphed relative to the brain decoding performance of the pre-trained BERT language model. Performance rapidly diverges and then stabilizes within tens of fine-tuning steps. Shaded regions represent 95\% confidence intervals, pooling across 8 subjects and up to 8 fine-tuning runs per model.}
    \label{fig:decoding-dynamics}
\end{figure*}

When during training do these models diverge in brain decoding performance? We repeat our brain decoding evaluation on model snapshots taken every 5 steps during fine-tuning, and chart brain decoding performance over time for each model in \Cref{fig:decoding-dynamics}. We find that models rapidly diverge in brain decoding performance, but remain mostly stable after about 100 fine-tuning steps. This phase of rapid change in brain decoding performance is generally matched with a phase of rapid change in task performance (compare each line in \Cref{fig:decoding-dynamics} with the learning curves in \Cref{fig:learning-curves}).

\subsection{Representational analysis}

We next investigate the structure of the model representations, and find that differing fidelity of \emph{syntactic} representation can explain some major qualitative differences between the models in brain decoding performance.

\subsubsection{Representational similarity}
\label{sec:rsa}
\newcommand\modeldist{\ensuremath{D_{j\ell}}}

\begin{figure}[t]
    \centering
    \includegraphics[width=0.9\linewidth]{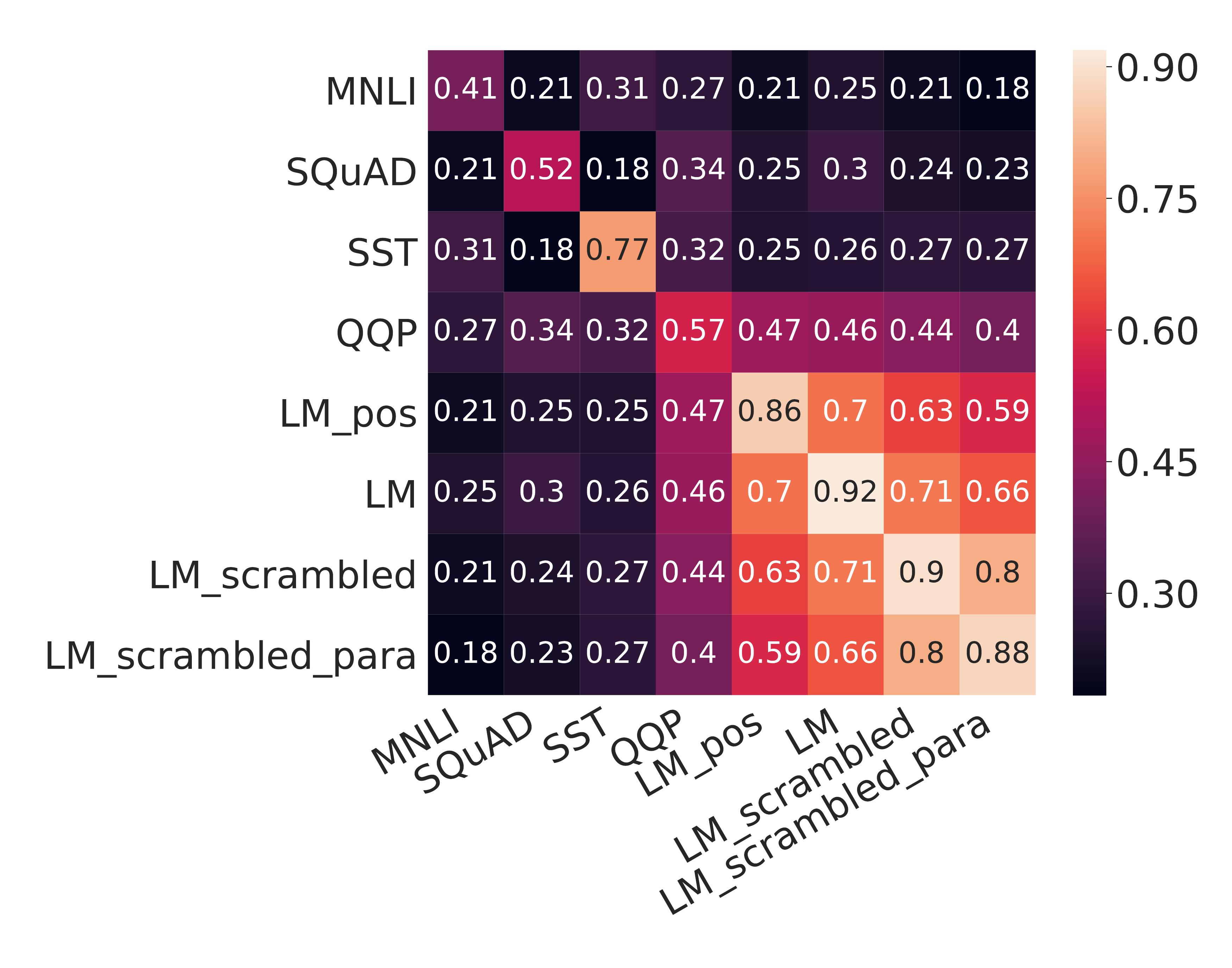}
    \caption{Representational similarity of the sentence encodings produced by each model (between -1 and 1; higher is more similar). See \Cref{sec:rsa} for details.}
    \label{fig:rsa}
\end{figure}

We first investigate coarse-grained model similarity with representational similarity analysis \citep{kriegeskorte2008representational}, which measures models' pairwise distance judgments as a proxy for how well the underlying representations are aligned in their contents.
For each fine-tuning run $\ell$ on each task $j$, we compute pairwise cosine similarities between each pair of sentence representation rows in $\cf$, yielding a vector $\modeldist \in \mathbb R^{(\npereira{} \text{ choose } 2)}$.

We measure the similarity between the representations derived from a run $(j,\ell)$ and some other run $(j',\ell')$ by computing the Spearman correlation coefficient $\rho(\modeldist, D_{j'\ell'})$. These correlation values are graphed as a heatmap in \Cref{fig:rsa}, where each cell averages over different runs of the two corresponding models. This heatmap yields several clear findings:

\begin{enumerate}
    \item The language modeling fine-tuning runs (especially the two LM-scrambled tasks) are the only models which have reliable high correlations between one another.
    \item 
    Language modeling tasks yield representations which make similar sentence-sentence distance predictions between different runs on the same task, while the rest of the models are less coherent across runs (see matrix diagonal).
\end{enumerate}

The scrambled LM tasks produce sentence representations which are reliably coherent across runs (\Cref{fig:rsa}), and produce reliable improvements in brain decoding performance (\Cref{fig:decoding-final}).
What is it about this task which yields such reliable results? We attempt to answer this question in the following section.

\subsubsection{Syntactic probe}
\label{sec:syntactic-probe}

\begin{figure}[t]
    \centering
    \includegraphics[width=\linewidth]{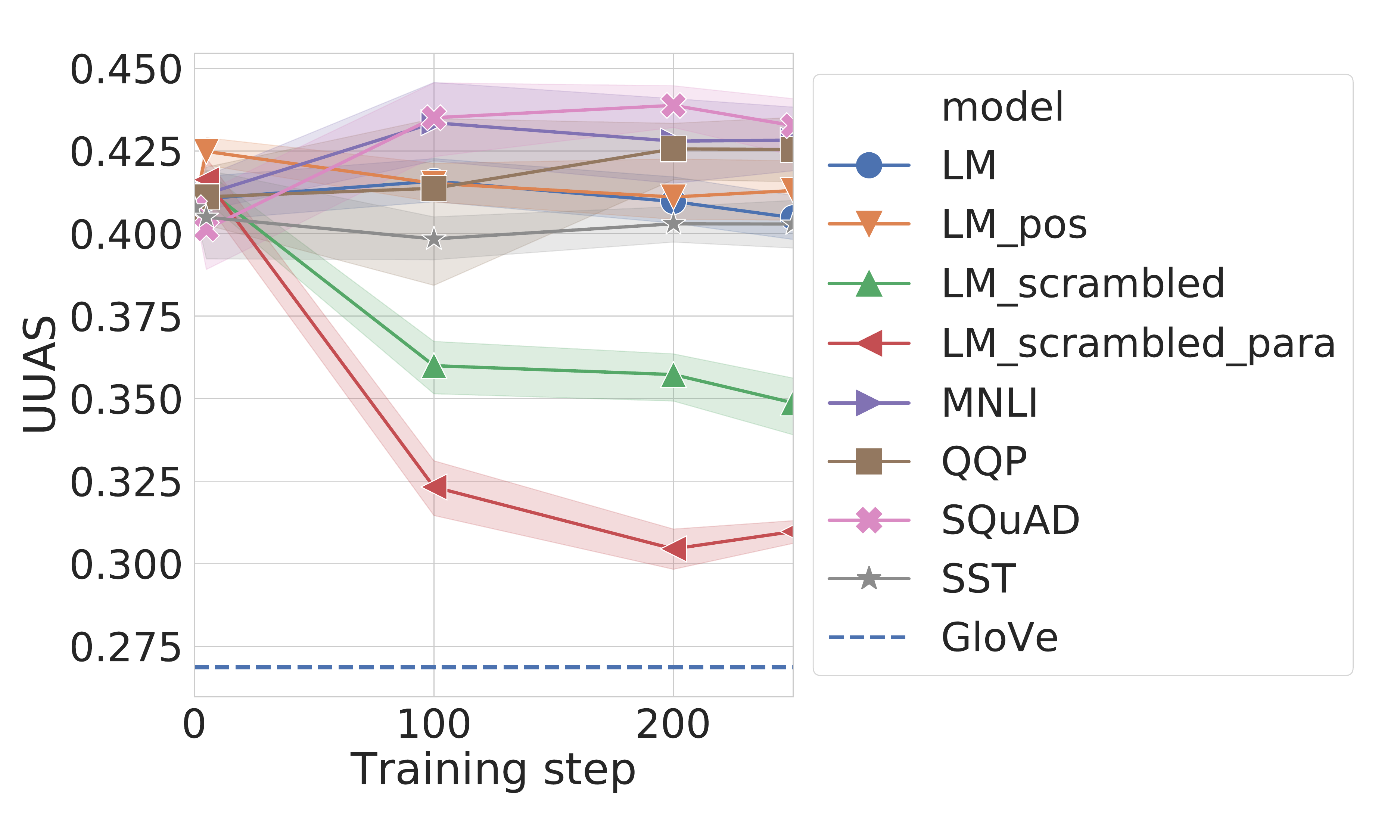}
    \caption{Syntactic probe evaluation across fine-tuning time (\cref{sec:syntactic-probe}).}
    \label{fig:syntactic-probe}
\end{figure}

\pgfkeys{%
/depgraph/reserved/edge style/.style = {%
black, solid, line cap=round, 
rounded corners=2, 
},%
}
\begin{figure}[t]
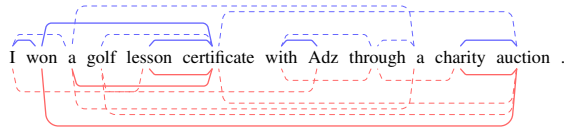

\centering
\resizebox{\linewidth}{!}{
\begin{dependency}[hide label, edge unit distance=0.7ex]
    \begin{deptext}[column sep=0.05cm]
    I\& won\& a\& golf\& lesson\& certificate\& with\& Adz\& through\& a\& charity\& auction\& . \\
\end{deptext}
\depedge[dashed, edge style={blue!60!}]{10}{3}{.} 
\depedge[thick, edge style={blue!60!}]{6}{5}{.} 
\depedge[dashed, edge style={blue!60!}]{12}{6}{.} 
\depedge[dashed, edge style={blue!60!}]{9}{7}{.} 
\depedge[thick, edge style={blue!60!}]{12}{11}{.} 
\depedge[thick, edge style={blue!60!}]{6}{2}{.} 
\depedge[thick, edge style={blue!60!}]{2}{1}{.} 
\depedge[dashed, edge style={blue!60!}]{10}{9}{.} 
\depedge[dashed, edge style={blue!60!}]{3}{1}{.} 
\depedge[thick, edge style={blue!60!}]{8}{7}{.} 
\depedge[dashed, edge style={blue!60!}]{6}{4}{.} 
\depedge[dashed, edge style={red!60!}, edge below]{10}{3}{} 
\depedge[thick, edge style={red!60!}, edge below]{6}{3}{} 
\depedge[thick, edge style={red!60!}, edge below]{12}{11}{} 
\depedge[thick, edge style={red!60!}, edge below]{6}{5}{.} 
\depedge[thick, edge style={red!60!}, edge below]{12}{2}{.} 
\depedge[dashed, edge style={red!60!}, edge below]{12}{4}{.} 
\depedge[dashed, edge style={red!60!}, edge below]{12}{6}{.} 
\depedge[dashed, edge style={red!60!}, edge below]{11}{9}{.} 
\depedge[dashed, edge style={red!60!}, edge below]{9}{7}{.} 
\depedge[dashed, edge style={red!60!}, edge below]{5}{1}{.} 
\depedge[dashed, edge style={red!60!}, edge below]{8}{4}{.} 
\end{dependency}
}
\caption{Example sentence with undirected syntactic probe parses induced from LM-scrambled representations (blue, above) and GloVe representations (red, below). Solid arcs represent correct dependency predictions; dashed lines represent errors.}
\label{fig:syntactic-probe-parse}
\end{figure}

Because the scrambled LM tasks were designed to remove all first-order cues to syntactic constituency from the input, we hypothesized that the models trained on this task were succeeding due to their resulting coarse syntactic representations. We tested this idea using the structural probe method of \citet{hewitt2019structural}. This method measures the degree to which word representations can reproduce syntactic analyses of sentences. We used dependency-parsed sentences from the Universal Dependencies (UD) English Web Treebank corpus \citep{silveira14gold} to evaluate the performance of each fine-tuned BERT model with a structural probe.

For each fine-tuned BERT model (task $j$, fine-tuning run $\ell$) and each sentence $w_1,\ldots,w_T$, let $\tilde{w_i}$ denote the context-sensitive representation of word $w_i$ under the model parameters $\thetaf$. The structural probe method attempts to derive a distance measure between context-sensitive word representations, parameterized by a transformation matrix $B$, which approximates the number of grammatical dependencies separating the two words \citep{hewitt2019structural}. Concretely, for any two words $w_i, w_j$ in a parsed sentence, we learn a parameter matrix $B$ such that
\begin{equation}
    \left((B(\tilde{w_i} - \tilde{w_j}))^T (B(\tilde{w_i} - \tilde{w_j}))\right)^2 \approx |w_i \leftrightarrow w_j|
\end{equation}
where $|w_i \leftrightarrow w_j|$ denotes the number of edges separating $w_i$ and $w_j$ in a dependency parse of the sentence.

We learn this parameter matrix $B$ for a set of training sentences randomly sampled from the UD corpus, and then apply the distance measure above to model representations for a set of held-out test sentences. For any sentence $w_1,\ldots,w_T$, the measure induces a $T \times T$ pairwise distance matrix, where each entry $(i,j)$ predicts the distance (in grammatical dependencies) between words $i$ and $j$. By applying a minimum spanning tree algorithm to this matrix, we derive an (undirected) parse tree for the sentence which best matches the predictions of the distance measure. We measure the accuracy of the reconstructed tree by calculating its unlabeled attachment score (UAS) relative to ground-truth parses from the UD corpus.

We apply the probe described above to every fine-tuning run of each model, and to baseline GloVe representations. We expected the GloVe representations to perform worst, since they cannot encode any context-sensitive features of input words.
The probe results are graphed over fine-tuning time in \Cref{fig:syntactic-probe}, relative to a probe induced from the GloVe representations (dashed blue line). This analysis shows that the models optimized for LM-scrambled and LM-scrambled-para --- the models which improve in brain decoding performance --- progressively worsen under this syntactic probe measure during fine-tuning. Their probe performance remains well above the performance of the GloVe baseline, however.

\Cref{fig:syntactic-probe-parse} shows a representative sample sentence with parses induced from the syntactic probes of LM-scrambled (after 250 fine-tuning steps) and the GloVe baseline. While both parses make many mistaken attachments (dashed arcs), the parse induced from LM-scrambled (blue arcs) makes better guesses about local attachment decisions than the parse from GloVe (red arcs), which seems to simply link identical and thematically related words. This is the case even though LM-scrambled is never able to exploit information about the relative positions of words during its training. Overall, this suggests that much (but not all) of the syntactic information initially represented in the baseline BERT model is \emph{discarded} during training on the scrambled language modeling tasks. Surprisingly, this loss of syntactic information seems to yield improved performance in brain decoding.

\section{Discussion}

The brain decoding paradigm presented in this paper has led us to a set of scrambled language modeling tasks which best match the structure of brain activations among the models tested. Optimizing for these scrambled LM tasks produces a rapid but stable divergence in representational contents, yielding improvements in brain decoding performance (\Cref{fig:decoding-final,fig:decoding-dynamics}) and reliably coherent predictions in pairwise sentence similarity (\Cref{fig:rsa}). These changes are matched with a clear loss of syntactic information (\Cref{fig:syntactic-probe}), though some minimal information about local grammatical dependencies is decodable from the model's context-sensitive word representations (\Cref{fig:syntactic-probe-parse}).

We do not take these results to indicate that human neural representations of language do not encode syntactic information. Rather, we see several possible explanations for these results:

\paragraph{Limitations of fMRI.} Functional magnetic resonance imaging --- the brain imaging method used to collect the dataset studied in this paper --- may be too temporally coarse to detect traces of the syntactic computations powering language understanding in the human brain.

This idea may conflict with several findings in the neuroscience of language. For example, \citet{brennan2016abstract} compared how language models with different granularities of syntactic representation map onto human brain activations during passive listening of English fiction. They derived word-level surprisal estimates from n-gram models (which have no explicit syntactic representation) and PCFG models (which explicitly represent syntactic constituency). In a stepwise regression analysis, they demonstrated that the surprisal estimates drawn from the PCFG model explain variance in fMRI measures of brain activation not already explained by estimates drawn from the n-gram model.

\citet{pallier2011cortical} examined a different hypothesis linking mental and neural representations of language. They presented subjects with strings of words which contain valid syntactic constituents of different lengths. They assumed that, since subjects will attempt to construct syntactic analyses of the word strings, the length of the possible syntactic constituents in any stimulus should have some correlate in subjects' neural activations. They found a reliable relationship between the size of the available constituents in the input and region-specific brain activations as measured by fMRI.

Our results are compatible with the idea that \emph{specific syntactic features} like those discussed above are still represented in the brain at the time scale of fMRI. \Cref{fig:syntactic-probe,fig:syntactic-probe-parse} demonstrate, in fact, that the \texttt{LM-scrambled} models still retain some syntactic information (or correlates thereof), in that they clearly outperform a baseline model in predicting the syntactic parses of sentences.

While these brain mapping studies have detected particular summary features of syntactic computation in the brain, these summary features do not constitute complete proposals of syntactic processing. In contrast, each of the models trained in this paper constitutes an independent candidate algorithmic description of sentence representation. These candidate descriptions can be probed (as in \Cref{sec:syntactic-probe}) to reveal exactly \emph{why} brain decoding fails or succeeds in any case.

Our paradigm thus enables us to next ask: what specific syntactic features are responsible for the improved performance of the \texttt{LM-scrambled} models? By further probing the models and designing ablated datasets, we plan to narrow down the particular phenomena responsible for the findings presented here. These results should act as a source of finer-grained hypotheses about what sort of syntactic information is preserved at coarse temporal resolutions, and allow us to resolve the conflict between our results those of \citet{pallier2011cortical} and \citet{brennan2016abstract}, among others.

\paragraph{Limitations of the linking hypothesis.} Our linear linking hypothesis (presented in \Cref{sec:methods-decoding}) that representations of syntactic structure are encoded entirely in the linear geometry of both neural networks and human brains. It is likely that some syntactic information --- among other features of the input --- \emph{are} conserved in the fMRI signal, but not readable by a linear decoder. Future work should investigate how more complex transforms linking brain and machine can reveal parallel structure between the two systems.

\paragraph{Limitations of the data.} The fMRI data used in this study (presented in \Cref{sec:methods-neuroimaging}) was collected as subjects read sentences and were asked to simply think about their meaning. This vague task specification may have led subjects to engage only superficially with the sentences. If this were the case, these shallow mental representations might present us with correspondingly shallow neural representations --- just the sort of representations which might be optimal for the simple tasks such as \texttt{LM-scrambled} and \texttt{LM-scrambled-para}. Future work should integrate brain images derived from different behavioral tasks, and study which model--brain relationships are conserved across these behaviors. Such studies could illuminate the degree to which there are genuinely task-general language representations in the mind.

\paragraph{}Our broader framework of analysis promises to reveal further insights about the parallel contents between artificial and human neural representations of language.
In this spirit, we have released our complete analysis framework as open source code for the research community, available at \url{http://bit.ly/nn-decoding}.

\section*{Acknowledgments}
JG gratefully acknowledges support from the Open Philanthropy Project, and RPL gratefully acknowledges support from a Newton Brain Science Research Seed Award and from the MIT--SenseTime Alliance on Artificial Intelligence.

\bibliography{main}
\bibliographystyle{acl_natbib}

\clearpage
\appendix

\section{Supplemental figures}
\label{sec:appendix-supp-fig}

\begin{figure*}[h]
\caption{Within-subject changes in brain decoding performance for different models, relative to the subject's brain decoding performance with the pre-trained BERT model. Error bars are 95\% CIs pooling across decoders learned for different runs of each model (up to 8 per model).}
\begin{subfigure}{\linewidth}
\includegraphics[width=\linewidth]{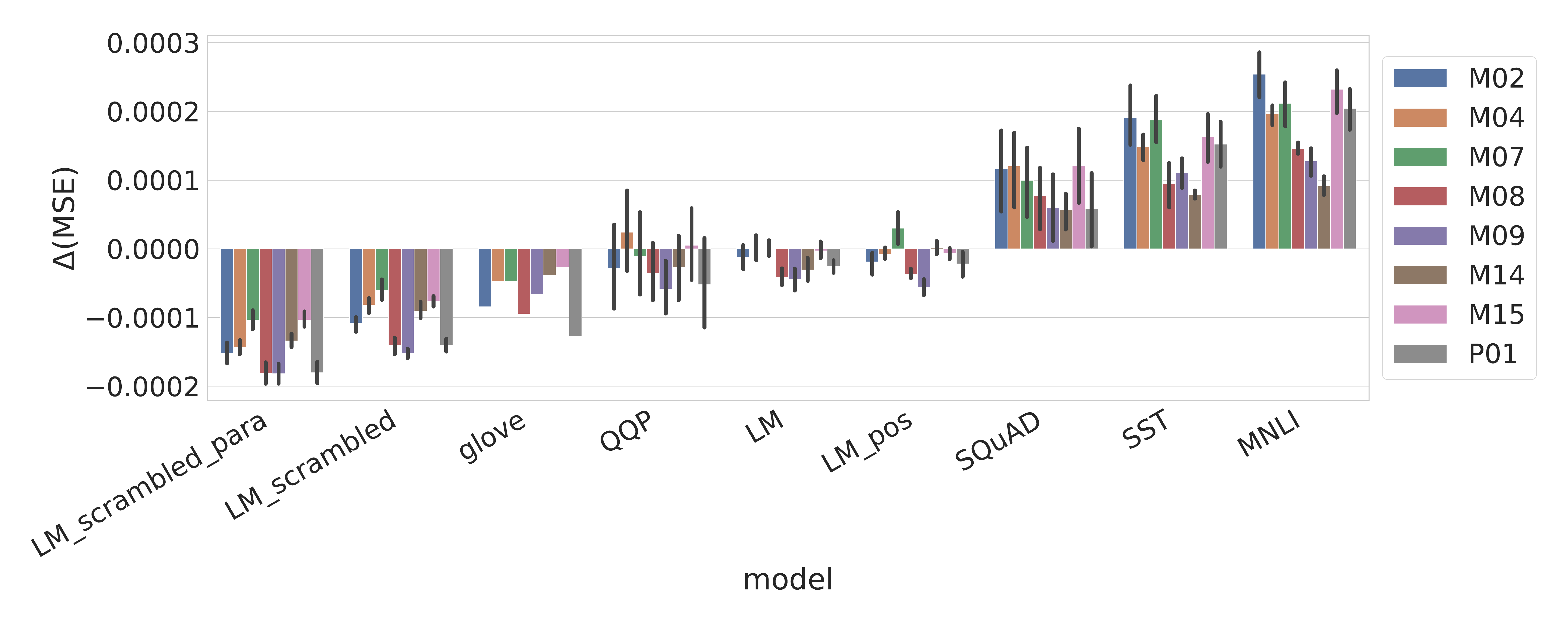}
\caption{Mean squared error metric}
\end{subfigure}\\[1em]
\begin{subfigure}{\linewidth}
\includegraphics[width=\linewidth]{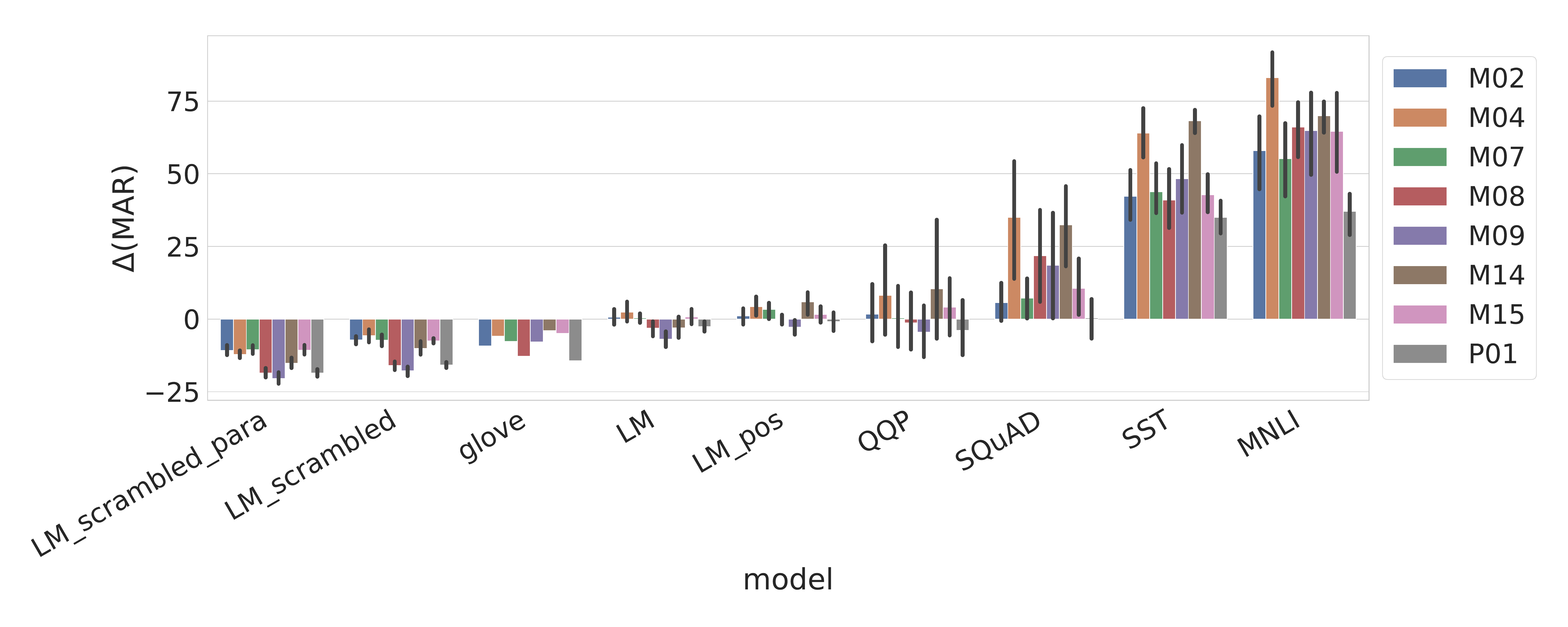}
\caption{Average rank metric}
\end{subfigure}
\label{fig:decoding-final-within-subject}
\end{figure*}

\Cref{fig:decoding-final-within-subject} shows the change in brain decoding performance after fine-tuning grouped by subject, under both the mean squared error and rank metrics.

\begin{figure*}[h]
\caption{Within-model changes in brain decoding performance for different subjects, relative to the corresponding subject's brain decoding performance with the pre-trained BERT model. Error bars are 95\% CIs pooling across decoders learned for different runs of each model (up to 8 per model).}
\begin{subfigure}{\linewidth}
\includegraphics[width=\linewidth]{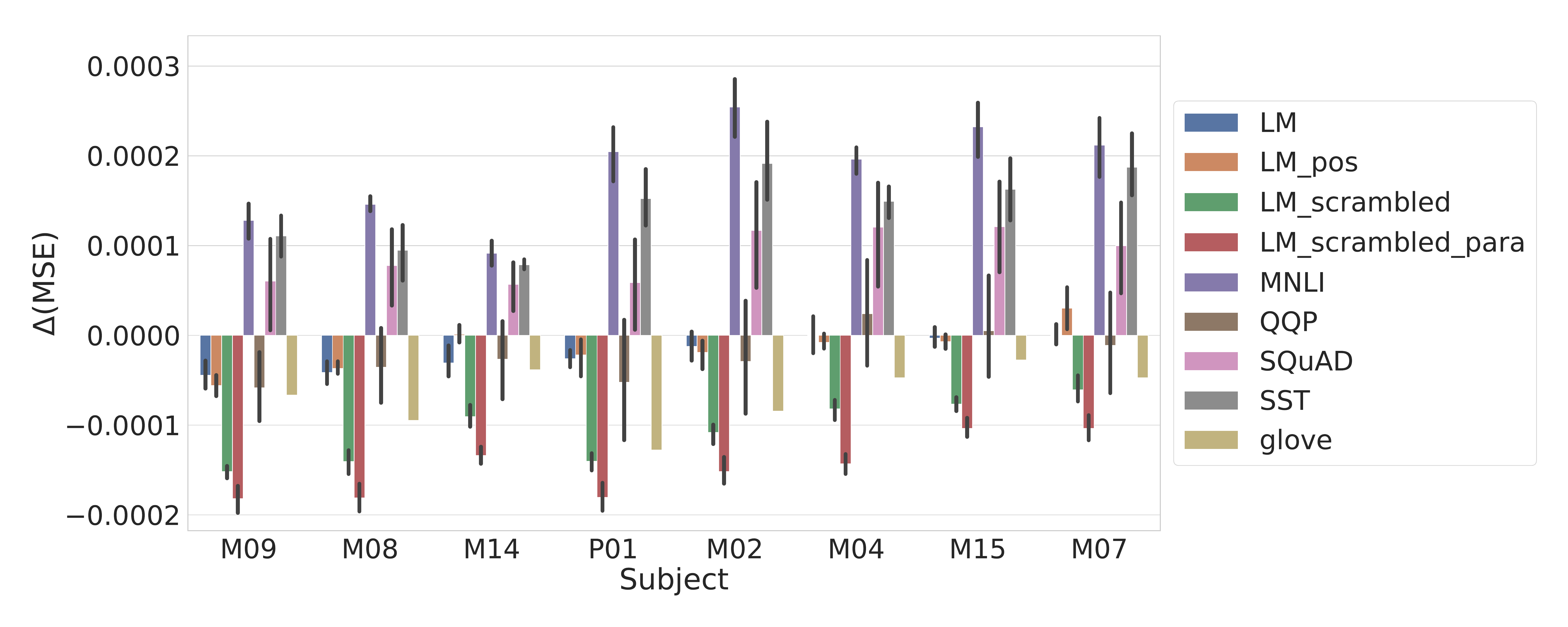}
\caption{Mean squared error metric}
\end{subfigure}\\[1em]
\begin{subfigure}{\linewidth}
\includegraphics[width=\linewidth]{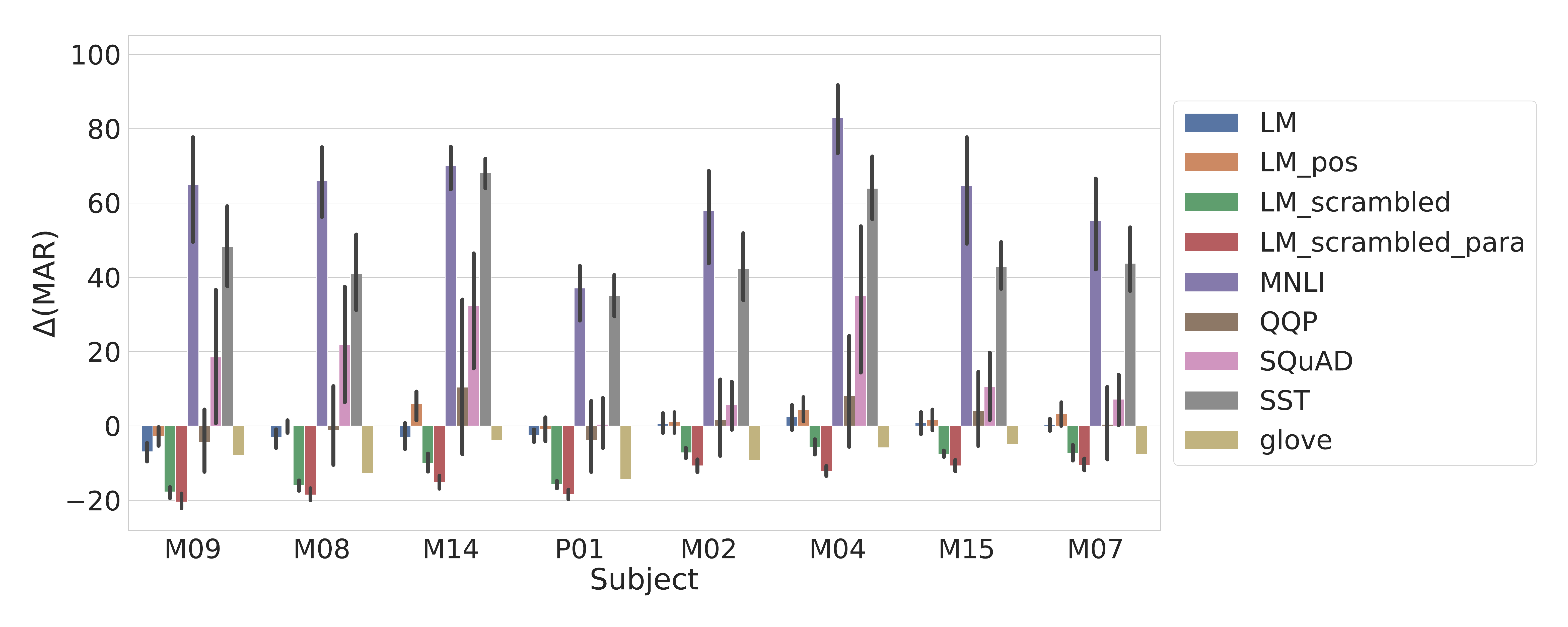}
\caption{Average rank metric}
\end{subfigure}
\label{fig:decoding-final-within-model}
\end{figure*}

\Cref{fig:decoding-final-within-model} shows the change in brain decoding performance after fine-tuning grouped by model, under both the mean squared error and rank metrics.

\section{Hyperparameters}
\label{sec:appendix-hyperparameters}

\begin{table}[h]
    \centering
    \begin{tabular}{l|c}
        \toprule
        Hyperparameter & Value  \\
        \midrule
        Batch size & 32 \\
        Learning rate & $2 \times 10^{-5}$\\
        LR warmup proportion & 10\% \\
        Maximum sequence length & 128 \\
        Fine-tuning steps & 250 \\
        \bottomrule
    \end{tabular}
    \caption{Fine-tuning hyperparameters, shared across fine-tuning runs for all tasks. These hyperparameters mostly follow the suggestions of \citet{devlin2018bert}.}
    \label{tbl:hyperparameters}
\end{table}

\begin{table}[h]
    \centering
    \begin{tabular}{l|c}
        \toprule
        Hyperparameter & Value  \\
        \midrule
        Training epochs & 10 \\
        Loss metric & L1 \\
        Maximum rank of $B$ & 30 \\
        Positive semi-definite? & Yes \\
        \bottomrule
    \end{tabular}
    \caption{Syntactic probe hyperparameters (\Cref{sec:syntactic-probe}), following the defaults of \citet{hewitt2019structural}.}
    \label{tbl:syntactic-probe-hyperparameters}
\end{table}

\section{Custom task information}
\label{sec:appendix-custom-tasks}

For each language modeling task, we randomly sampled and concatenated documents from the Toronto Books Corpus \citep{kiros2015skip}. Each language modeling dataset contained 1,000,000 training sentences and 100,000 development and test sentences. (We generated over-sized datasets in order to ensure that multiple runs of the same model would be highly unlikely to see similar samples of training data.)

For the part-of-speech task, we tagged each sentence using spaCy \citep{spacy2} and followed the same random masking procedure as in the typical cloze language modeling task. spaCy assigned 49 unique part-of-speech tags to the sentences, yielding a 49-way classification task.

For all tasks, we retained the secondary BERT objective requiring the model to predict whether two sentences are adjacent or not in a source document. (This objective did not differ from the standard setup for the part-of-speech task; for the scrambling task, the input sentences were independently randomly shuffled.)

\begin{figure*}[tbh]
\caption{Learning curves (training set loss and evaluation set accuracy) for custom language modeling tasks.}
\label{fig:learning-curves-lm}
\begin{subfigure}{0.24\linewidth}
\includegraphics[width=\linewidth]{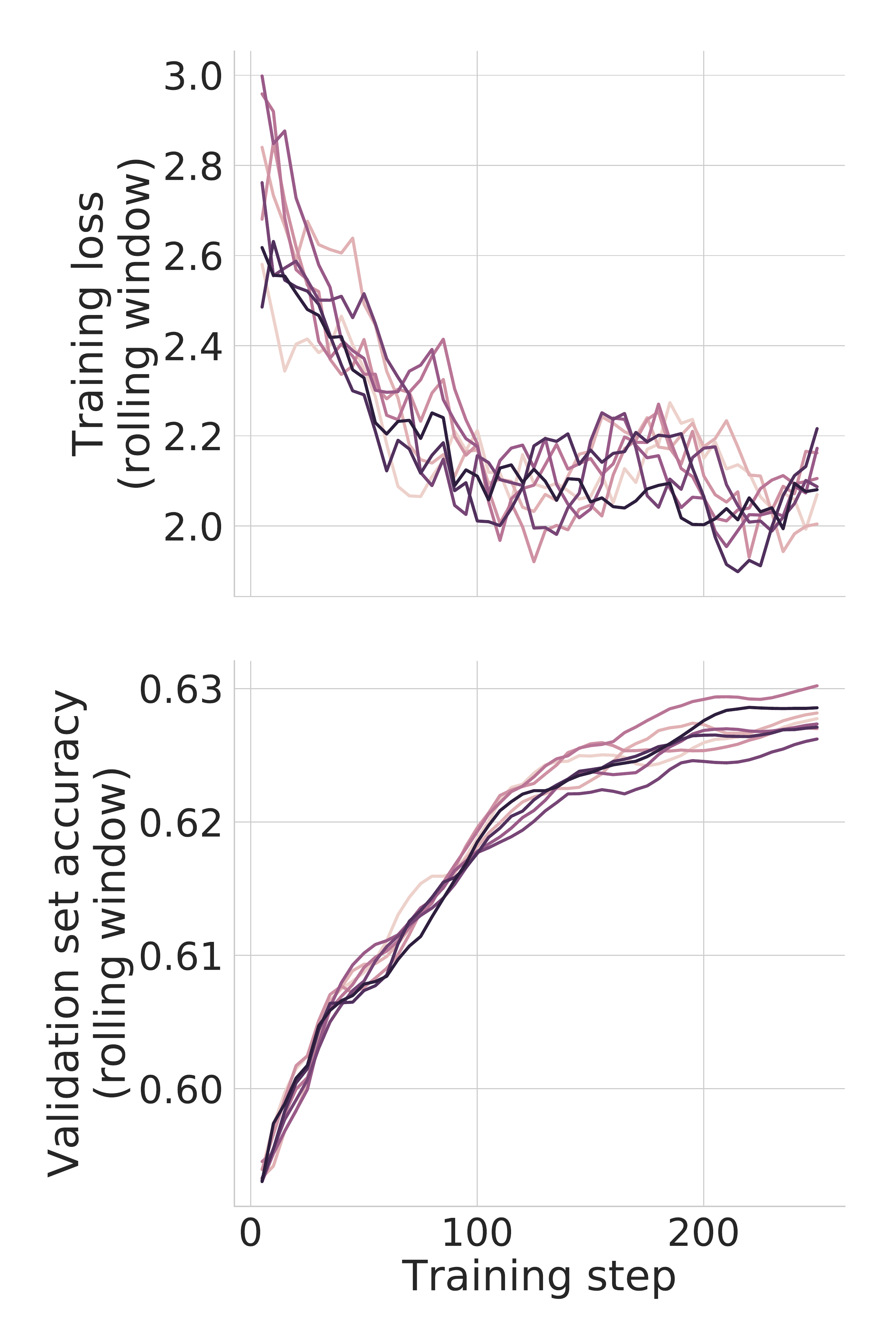}
\caption{LM (standard BERT task). Chance accuracy: $\frac{1}{30522} \approx 3 \times 10^{-5}$.}
\end{subfigure}\hfill%
\begin{subfigure}{0.24\linewidth}
\includegraphics[width=\linewidth]{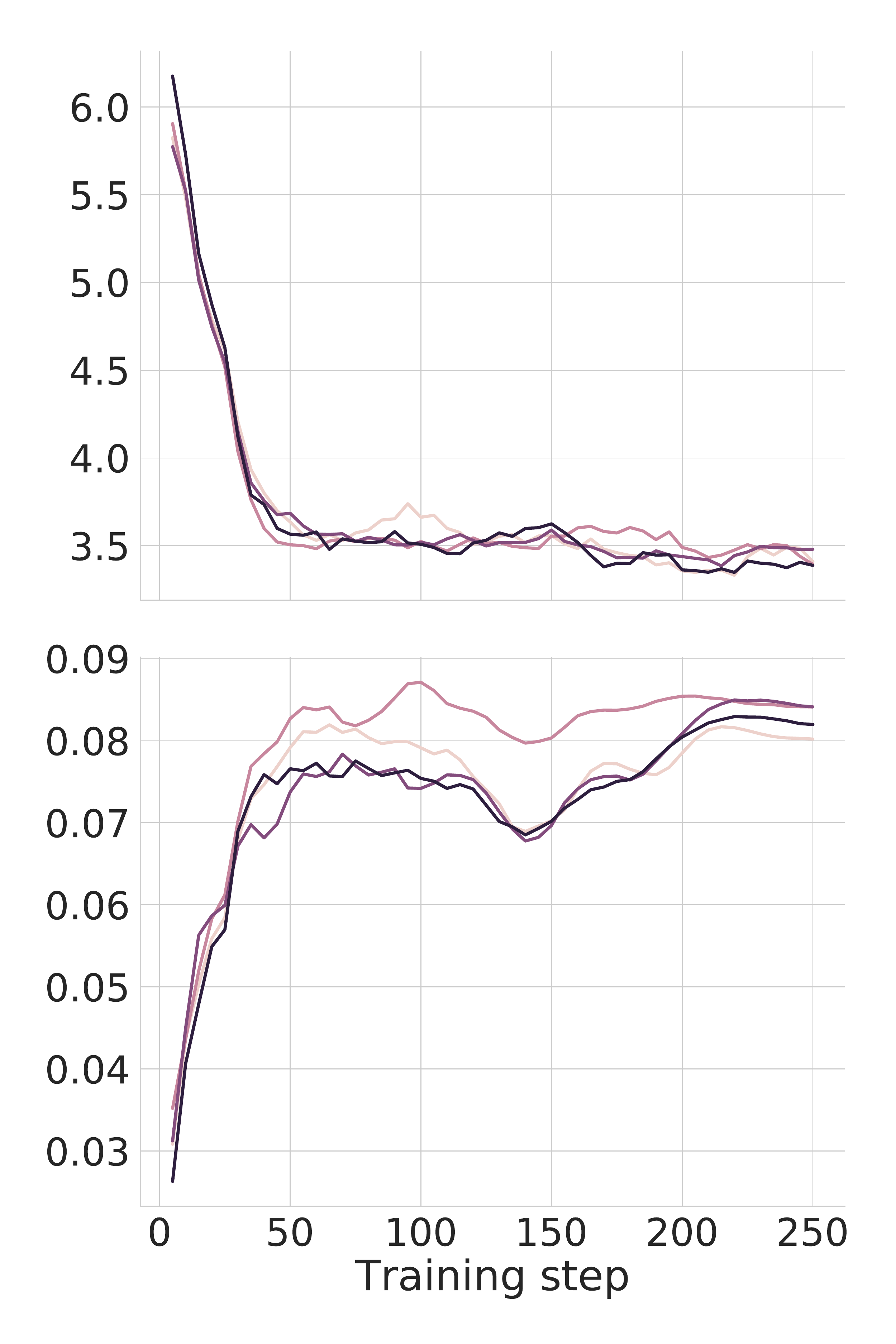}
\caption{LM-pos. Chance accuracy: $\frac{1}{49} \approx 0.02$.}
\end{subfigure}\hfill%
\begin{subfigure}{0.24\linewidth}
\includegraphics[width=\linewidth]{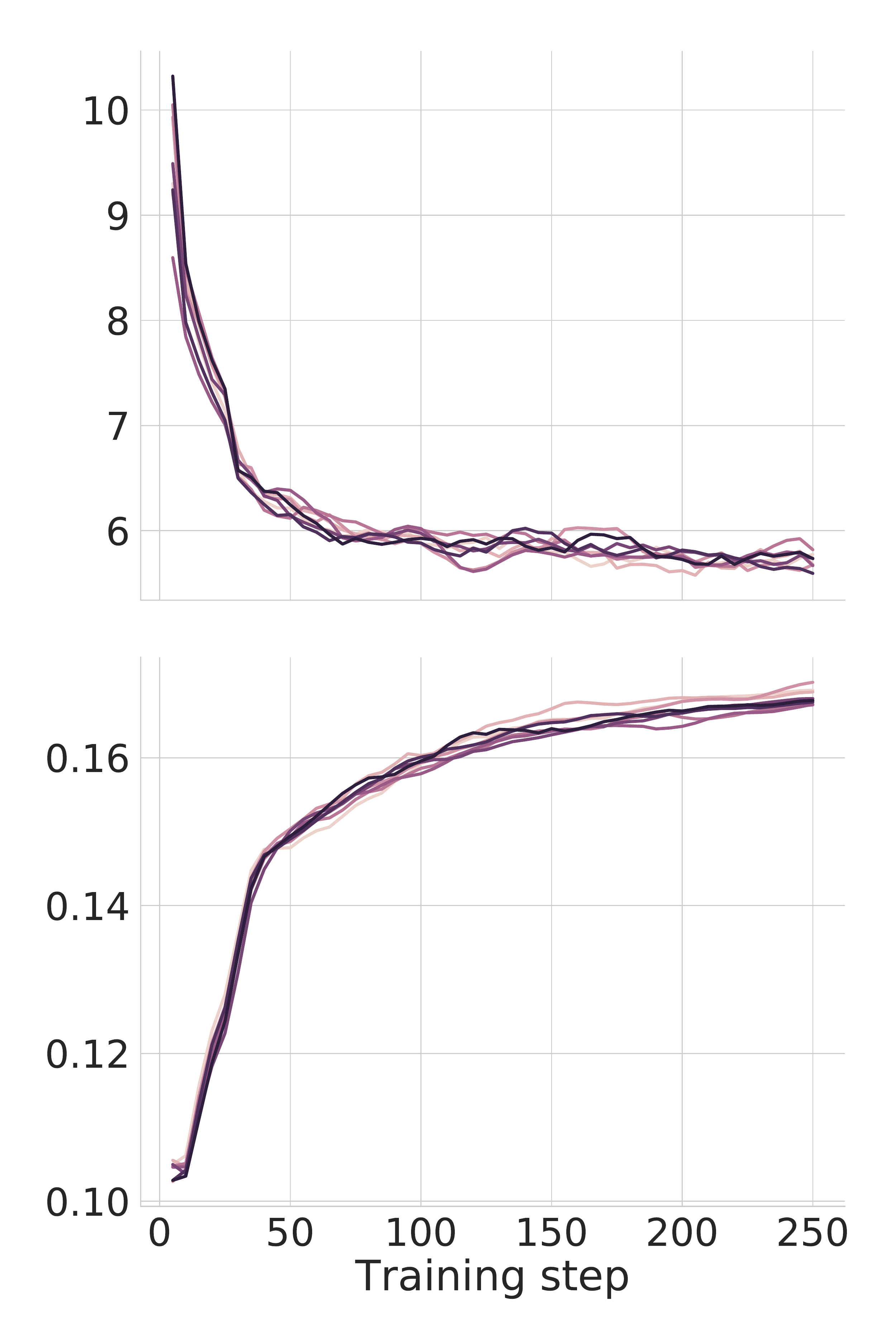}
\caption{LM-scrambled. Chance accuracy: $\frac{1}{30522} \approx 3 \times 10^{-5}$.}
\end{subfigure}\hfill%
\begin{subfigure}{0.24\linewidth}
\includegraphics[width=\linewidth]{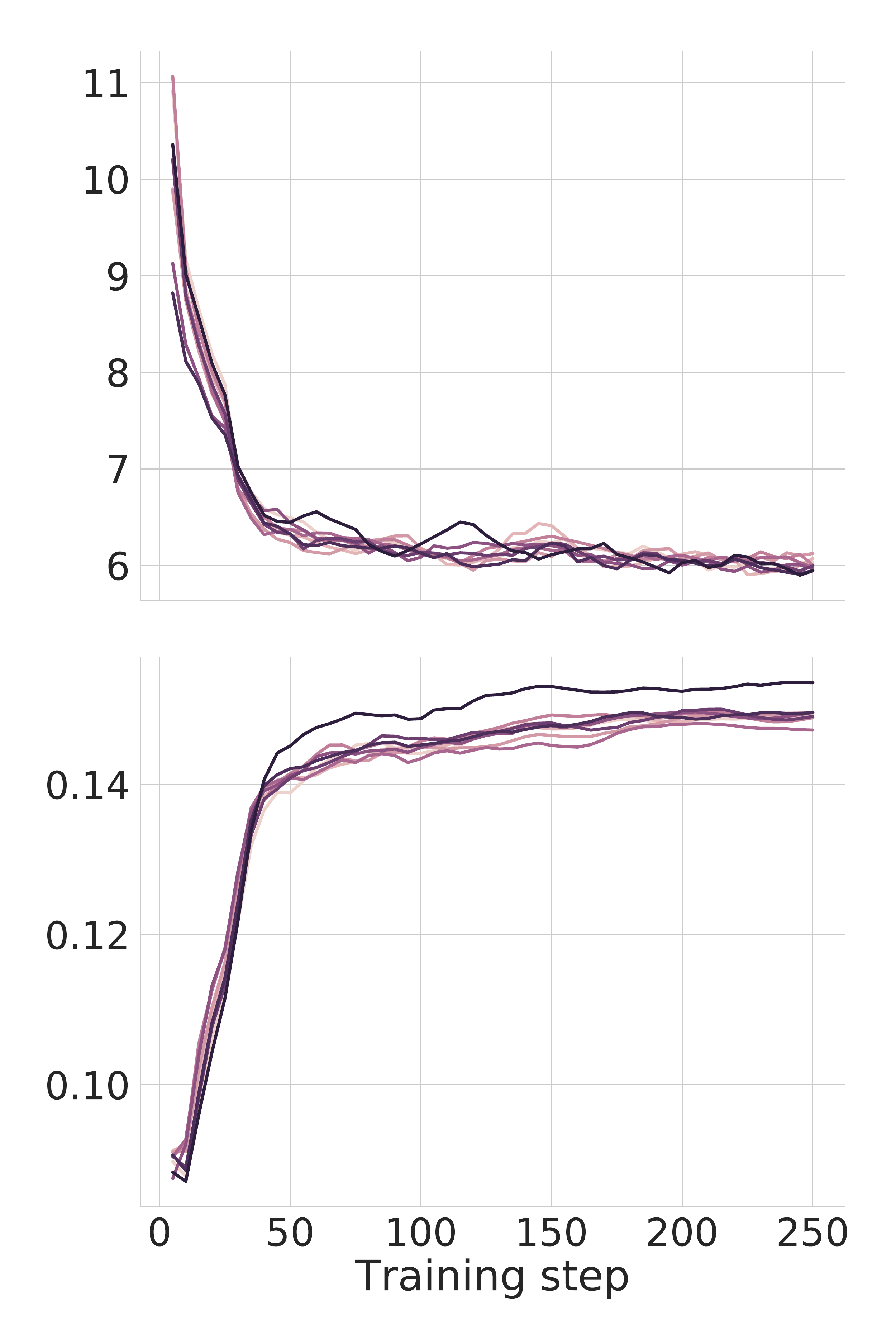}
\caption{LM-scrambled-para. Chance accuracy: $\frac{1}{30522} \approx 3 \times 10^{-5}$.}
\end{subfigure}
\end{figure*}

\Cref{tbl:examples-lm-scrambled} shows training examples from each of these custom tasks. \Cref{fig:learning-curves-lm} shows learning curves and validation accuracy curves for the models trained on each task.

\bgroup
\def\arraystretch{1.5}
\begin{table*}[tbh]
    \centering
    \caption{Examples from custom language modeling tasks.}
    \begin{subtable}{0.49\linewidth}
    \resizebox{\linewidth}{!}{
    \begin{tabular}{p{5cm}|l}
        \toprule
        Input & Ground-truth output \\
        \midrule{}
        [MASK] minutes in she began to cry . & two \\
        the door opened and lilith [MASK] in with worry on her face . & walked \\
        she grabbed several pairs of jeans and some t - shirts [MASK] her closet & from \\
        \bottomrule
    \end{tabular}
    }
    \caption{LM}
    \end{subtable}\hfill%
    \begin{subtable}{0.49\linewidth}
    \resizebox{\linewidth}{!}{
    \begin{tabular}{p{5cm}|l}
        \toprule
        Input & Ground-truth output \\
        \midrule
        , and [MASK] so drive can . & i \\
        instead \#\#k kn documents pots important dish water be \#\#s greasy , [MASK] were of of and soon legal in dirty \#\#ick personal \#\#ks high to a bath and looking pan awaiting the i with ago pan lifestyle face many \#\#ils d , years . & piles \\
        by connected will if you the \#\#tell we the that \#\#i [MASK] us lead ? & people \\
        , and i so drive can . abraham want , i myself i [MASK] cars and whenever his , two , dad & have \\
        \bottomrule
    \end{tabular}
    }
    \caption{LM-scrambled}
    \end{subtable}\\[1em]
    
    \begin{subtable}{0.49\linewidth}
    \resizebox{\linewidth}{!}{
    \begin{tabular}{p{5cm}|l}
        \toprule
        Input & Ground-truth output \\
        \midrule
        scent sweat ta very [MASK] aroma \#\#ed caught , , a close a his . \#\#ana cabaret sand male skin she \#\#wny of up and \#\#al mu \#\#ting slick faint ... & \#\#wood \\
        , no in ! themselves supper with eyes included his ` , [MASK] began . the the alone everyone cried enjoying standing ... & triumph \\
        , she told off out would wash followed `` asked her i with '' take . voice [MASK] ... & stepped \\
        \bottomrule
    \end{tabular}
    }
    \caption{LM-scrambled-para}
    \end{subtable}\hfill%
    \begin{subtable}{0.49\linewidth}
    \resizebox{\linewidth}{!}{
    \begin{tabular}{p{5cm}|l}
        \toprule
        Input & Ground-truth output \\
        \midrule
        you ' ve probably just gotten yourself off schedule a few days [MASK] you've been so busy and stressed . & \texttt{CD} \\
        a couple of lordlings [MASK] . & \texttt{VBD} \\
        the blond boy , who at fourteen was [MASK] much taller than anderra ... showed no signs of discomposure . & \texttt{RB} \\
        \bottomrule
    \end{tabular}
    }
    \caption{LM-pos}
    \end{subtable}
    \label{tbl:examples-lm-scrambled}
\end{table*}
\egroup

\end{document}